\documentclass[lettersize,journal]{IEEEtran}
\usepackage{cite}
\usepackage{url}
\usepackage{ragged2e}
\usepackage{footnote}
\makesavenoteenv{tabular}
\makesavenoteenv{table}
\usepackage{epsfig}
\usepackage{graphicx}
\usepackage{amsmath,amssymb} % define this before the line numbering.
\usepackage{algorithm}
\usepackage{algorithmicx}
\usepackage{algpseudocode}
\usepackage{graphics}
\usepackage{threeparttable}
\usepackage{color}
\usepackage[normalem]{ulem}
\usepackage{multirow}
\usepackage{float}
\usepackage{amsfonts}
\usepackage{bm}
\usepackage{array}
\usepackage{pifont}
\usepackage{diagbox}
\usepackage{rotating}
\usepackage{booktabs}
\usepackage{overpic}
\usepackage{textcomp}
\usepackage{contour}

\usepackage{enumitem}
\usepackage{colortbl}
\usepackage{csquotes}
\usepackage[american]{babel}
\usepackage{microtype}
\usepackage{bbding}
\usepackage{amsfonts,amssymb}
\usepackage[pagebackref=false,breaklinks=true,colorlinks, bookmarks=false]{hyperref}
\usepackage[table,xcdraw]{xcolor}
\usepackage{colortbl}
\usepackage{cleveref}
\crefformat{section}{\S#2#1#3} 
\crefformat{subsection}{\S#2#1#3}
\crefformat{subsubsection}{\S#2#1#3}
\usepackage{silence}
\usepackage{booktabs}
\usepackage{makecell}
\def\ie{\emph{i.e.}}
\def\eg{\emph{e.g.}}

\def\etal{{\em et al.~}}

\graphicspath{{./Imgs/}{../../figure/}{./Imgs/results/}{./Imgs/authors/}}

\ifdefined \GramaCheck
  \newcommand{\CheckRmv}[1]{}
  \renewcommand{\eqref}[1]{Equation 1}
\else
  \newcommand{\CheckRmv}[1]{#1}
  \renewcommand{\eqref}[1]{Equation~(\ref{#1})}
\fi

\begin{document}

\title{Semi-Cycled Generative Adversarial Networks for Real-World Face Super-Resolution}

\author{Hao~Hou,
        Jun~Xu,~\IEEEmembership{Member,~IEEE,}
        Yingkun~Hou,~\IEEEmembership{Senior~Member,~IEEE,}
        Xiaotao~Hu,
        Benzheng~Wei,
        and
        Dinggang~Shen,~\IEEEmembership{Fellow,~IEEE}
\IEEEcompsocitemizethanks{
\IEEEcompsocthanksitem
This research is supported in part by The National Natural Science Foundation of China (No. 62002176, 62176068, and 61872225), Natural Science Foundation of Shandong Province (No. ZR2020MF038, ZR2020KF013, and ZR2020ZD44).
Corresponding author: Jun Xu (nankaimathxujun@gmail.com)
\IEEEcompsocthanksitem H. Hou is with College of Intelligence and Information Engineering, Shandong University of Traditional Chinese Medicine, Ji'nan, 250355, China.
\IEEEcompsocthanksitem J. Xu is with School of Statistics and Data Science, KLMDASR, LEBPS, and LPMC, Nankai University, Tianjin 300071, China, and with School of Biomedical Engineering, ShanghaiTech University, Shanghai, China.
\IEEEcompsocthanksitem Y.-K. Hou is with the School of Information Science and Technology, Taishan University, Tai'an, 271000, China.
\IEEEcompsocthanksitem X.-T. Hu is with College of Computer Science, Nankai University, Tianjin 300071, China.
\IEEEcompsocthanksitem B.-Z. Wei and H. Hou are with the Center for Medical Artificial Intelligence, Shandong University of Traditional Chinese Medicine, Ji'nan, 250355, China.
\IEEEcompsocthanksitem D.-G. Shen is with School of Biomedical Engineering, ShanghaiTech University, and United Imaging Co., Shanghai, China.}}

%\markboth{Journal of \LaTeX\ Class Files,~Vol.~14, No.~8, August~2021}

% % column for its text to clear the IEEEpubid mark.

\maketitle

\begin{abstract}
Real-world face super-resolution (SR) is a highly ill-posed image restoration task. The fully-cycled Cycle-GAN architecture is widely employed to achieve promising performance on face SR, but is prone to produce artifacts upon challenging cases in real-world scenarios, since joint participation in the same degradation branch will impact final performance due to huge domain gap between real-world~and synthetic LR ones obtained by generators. To better exploit the powerful generative capability of GAN for real-world face SR, in this paper, we establish two independent degradation branches in the forward and backward cycle-consistent reconstruction processes, respectively, while the two processes share the same restoration branch. Our Semi-Cycled Generative Adversarial Networks (SCGAN) is able to alleviate the adverse effects of the domain gap between the real-world LR face images and the synthetic LR ones, and to achieve accurate and robust face SR performance by the shared restoration branch regularized by both the forward and backward cycle-consistent learning processes. Experiments on two synthetic and two real-world datasets demonstrate that, our SCGAN outperforms the state-of-the-art methods on recovering the face structures/details and quantitative metrics for real-world face SR. The code will be publicly released at \url{https://github.com/HaoHou-98/SCGAN}.
\end{abstract}


\vspace{-3mm}

\section{Introduction}
\IEEEPARstart{F}{ace} is of central importance for human identity recognition.
The low-resolution (LR) face images captured by camera sensors would largely degrade the corresponding identity information.
Face super-resolution (SR) aims to estimate high-resolution (HR) face images from LR ones, to improve the image quality and performance of subsequent identity recognition tasks~\cite{FASEL2003,zhao2003face,xiong2013supervised}.
This task is very challenging upon complex real-world scenarios, where the degradation kernel is usually unknown.
Traditional face SR methods can be roughly divided into local patch-based methods~\cite{liu2005neighbor,park2007robust,chang2004super}, global image-based methods~\cite{wang2005hallucinating,chakrabarti2007super,park2008example}, and hybrid methods taking advantage of global image consistency and local patch sparsity~\cite{hu2010learning,hu2010local,li2014face,shi2014global}.
% 缺乏特性
However, these hand-crafted methods could hardly achieve satisfactory results upon diverse real-world scenarios~\cite{gpen_cvpr2021}.

\CheckRmv{
\begin{figure}[t] % htbp
	\centering
    \small
	\begin{overpic}[width=\linewidth]{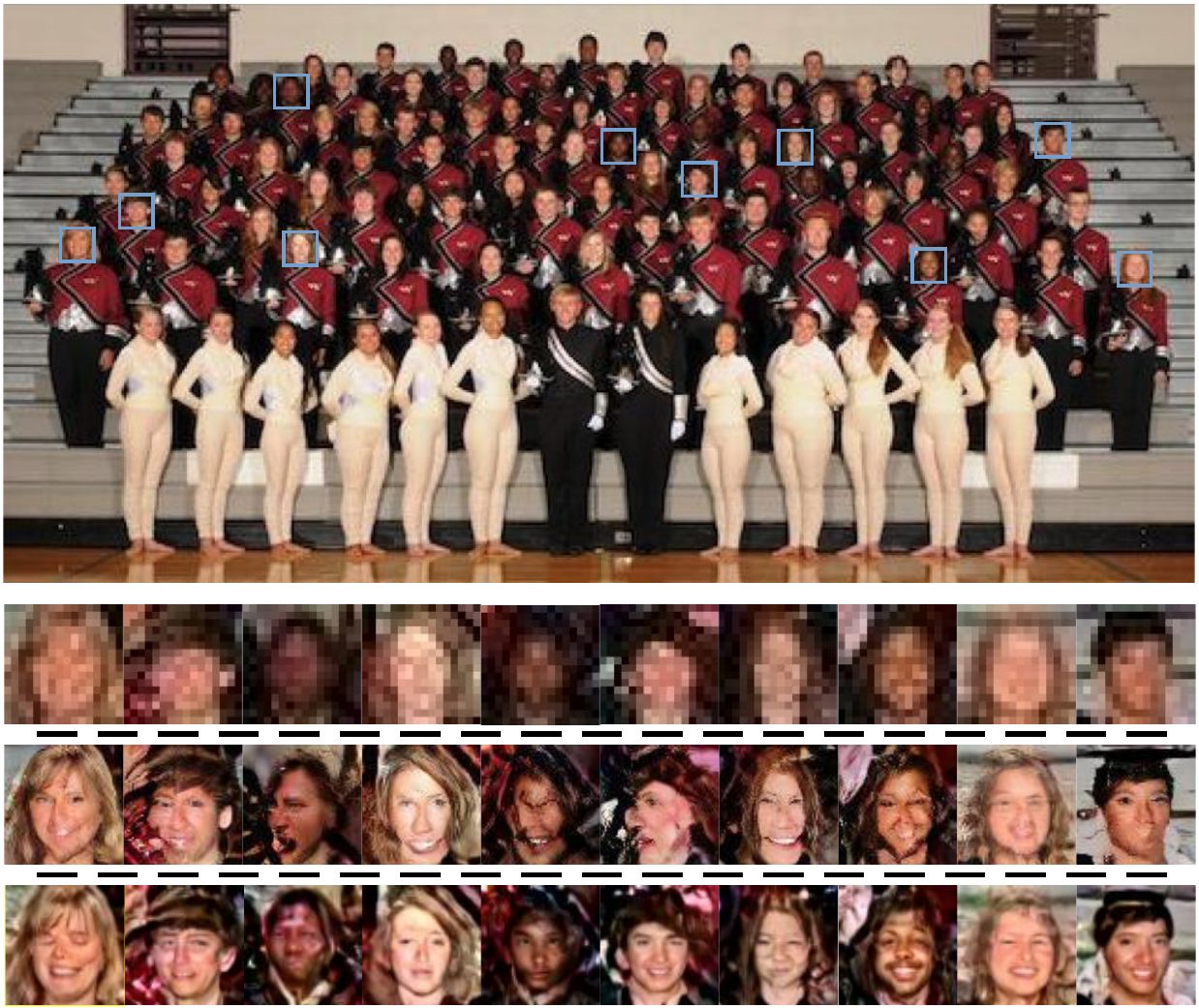}
    \end{overpic}
    \vspace{-6mm}
	\caption{\textbf{Comparison of LRGAN~\cite{LRGAN} and our SCGAN on blind real-world face SR}.
	% face SR results by
	We perform $\times4$ real-world face SR on $16\times16$ LR face images to obtain $64\times64$ HR ones.
	1-st row: a real-world group photo crawled from the internet that suffers from complex and unknown degradation.
	2-nd row: the LR face images from the photo.
	3-rd row: the face SR results of LRGAN~\cite{LRGAN}.
	4-th row: the SR results by our SCGAN.
	% \textbf{Please zoom in for better view}.
	}
    \label{fig:example}
    \vspace{-5mm}
\end{figure}
\vspace{0.01mm}
}

\par
Recently, the powerful learning capability of deep convolutional neural networks (CNNs) has been extensively exploited for face SR~\cite{huang2016face,liu2020densely,chen2020learning,chen2020rbpnet,huang2017wavelet}.
These discriminative CNNs mainly learn a direct enhancing mapping function between pairs of LR and HR face images.
For objective evaluation, the LR face images are usually degraded by synthetic downsampling kernels from the HR ones.
However, since it is difficult to obtain the corresponding HR face images for the real-world LR ones, the discriminative CNNs suffer from a huge performance gap between synthetic and practical degradation for real-world face SR.
To this end, several methods~\cite{GFRNet_eccv2018,GWAInet_cvprw2019,DFDNet_eccv2018} align LR face images with unpaired HR face images with the same identity.
However, face alignment is often challenged by the insufficiently trained face SR models, due to short of HR face images in practical scenarios.

\par
Compared to the discriminative competitors, generative CNNs like Generative Adversarial Networks (GANs)~\cite{GAN2014} are employed in~\cite{li2020enhanced,GFRNet_eccv2018,chen2021progressive,pulse_cvpr2020,GFPGAN2021,gpen_cvpr2021} to perform blind face SR with complex degradations.
To deal with unknown real-world degradation, several generative CNNs~\cite{LRGAN,maeda2020unpaired,guo2020closed,zhang2020deblurring} further implement unsupervised face SR by resorting to the insight of cycle-consistency developed for the unpaired image translation tasks~\cite{cyclegan2017}.
LRGAN~\cite{LRGAN} is a representative work to utilize the cycle learning scheme~\cite{cyclegan2017} for real-world face SR, introducing a ``learning-to-degrade'' branch and a ``learning-to-SR'' branch to perform face image degradation and SR, respectively.
However, since unpaired LR and HR face images suffer from a considerable gap on identity information, the two branches in LRGAN are consistent only for the HR face images and could hardly preserve well the face details and identity information of the LR face images.
\par
Since the unpaired LR and HR face images suffer from uncertain relationship, employing a directional framework~\cite{LRGAN} or a fully-cycled bidirectional one~\cite{cyclegan2017} is not sufficient to simultaneously preserve the identity information of the LR and HR face images in real-world scenarios.
To better alleviate the domain gap between unpaired LR and HR face images, in this paper, we introduce a Semi-Cycled Generative Adversarial Network (SCGAN) for real-world face SR, by extending the bidirectional cycle consistency scheme in~\cite{cyclegan2017} to a more flexible version.
Specifically, we propose to learn three generative branches, instead of two in~\cite{cyclegan2017,LRGAN}, for real-world HR and LR face image reconstructions: 1) a ``learning-to-degrade'' branch to obtain synthetic LR face images by degrading the HR ones, 2) a ``learning-to-SR'' branch to obtain the SR images by restoring the synthetic and real-world LR face images, and 3) another ``learning-to-degrade'' branch to degrade the SR images restored from the real-world LR images.
%  the three branches in 
Different from CycleGAN~\cite{cyclegan2017}, our SCGAN is only coupled at the middle ``learning-to-SR'' branch, while learning the cycle consistency of LR and HR face image reconstructions by individual branches.
% we propose to establish independent degradation mappings for the forward and backward processes, and then establish their degradation mappings, respectively, thus provide the restoration branch with training LR face images that more closely fits the real data distribution.
% The forward and backward processes share the same restoration branch to constrain each other, and the data diversity for training the restoration branch will be improved, thereby improving the generalization ability of the model. 
%
For example, in Figure~\ref{fig:example}, we compare the real-world face SR performance between LRGAN~\cite{LRGAN} and our SCGAN.
The real-world LR face images with severe degradation could hardly be restored by LRGAN to recover the identity structure and details.
However, our SCGAN, benefited from the semi-cycle consistency insight, well preserves both aspects for face SR.
\par 
In summary, our contributions are mainly three-fold:

\begin{itemize}
    \item \textbf{We develop a novel Semi-Cycled architecture to exploit GANs for real-world face super-resolution}.
    Our proposed Semi-Cycled GANs (SCGAN) well mitigate adverse effects of the degradation gap between real-world LR face images and synthetic ones, resulting in better preservation of identity and detailed information. 
    \item \textbf{We study in-depth the roles of adversarial loss, pixel loss, and cycle-consistency loss} in our SCGAN for face image super-resolution.
    The adversarial loss reduces the domain gap between the HR images and those outputs by our SCGAN, and the pixel loss enriches the contextual details of the SR results, while the cycle-consistency loss helps to preserve the structural information.
    \item Experiments on five benchmark datasets show that \textbf{our SCGAN outperforms the state-of-the-art methods quantitatively and qualitatively} on real-world face SR. Application on downstream vision tasks of face detection, face verification and face landmark detection further validates the effectiveness of our SCGAN on face SR. 
\end{itemize}

The rest of this paper is organized as follows.
In \S\ref{sec:relatedwork}, we summarize the related works.
In \S\ref{sec:method}, we introduce our SCGAN for real-world face SR.
In \S\ref{sec:experiments}, experiments on benchmark datasets are conducted to evaluate the performance of different face SR methods.
\S\ref{sec:conclusion} concludes this paper.
\vspace{-3mm}

\section{Related Work}\label{sec:relatedwork}

\subsection{Human Face Super-Resolution}
Human face super-resolution (SR) aims to obtain visual-pleasing high-resolution (HR) face images from the low-resolution ones~\cite{FSRsurvey_2021,maskguided_eccv2022}.
Early face SR methods~\cite{baker2000,gunturk_TIP2003,chang2004super,liu2005neighbor,park2007robust,LPH_PR2007} utilize hand-craft image priors and degradation models.
For instance, Baker \etal~\cite{baker2000} utilized Gaussian image pyramids for face SR, while Gunturk \etal~\cite{gunturk_TIP2003} presented a Bayesian model for face SR from a global image-level perspective.
% After that, Liu et al.~\cite{} proposed to combine the principal component analysis model with the markov random field model to deal with the face super-resolution task.
To well recover local details, the methods in~\cite{chang2004super,liu2005neighbor,park2007robust} tackle the face SR by patch-wise modeling.
Neighborhood embedding~\cite{chang2004super} is a representative work in this direction.
Later, the methods of~\cite{LPH_PR2007,hu2010learning,hu2010local,li2014face,shi2014global} have been developed for face SR to simultaneously preserve local details and global structures.
However, these methods do not perform well upon complex real-world cases.

Recent methods~\cite{huang2016face,liu2020densely,chen2020learning,chen2020rbpnet,huang2017wavelet} employ deep convolutional neural networks (CNNs) for face SR.
RBPNet~\cite{chen2020rbpnet} employs iterative back projection to directly learn the mapping from LR to HR face images.
SPARNet~\cite{chen2020learning} integrates the spatial attention mechanism into their framework to improve the representation ability of the network.
WaSRNet~\cite{huang2017wavelet} transforms the face image domain into the wavelet coefficient domain to preserve more details.
Lu et al.~\cite{lu2020global} proposed a hybrid approach based on a global upsampling network and a local enhancement network to jointly enhance the facial contours and local details.
The Residual Attribute Attention Network~\cite{RAAN_AAAI2019} employs a multi-block cascaded structure to extract pixel-level representation and semantic-level identification information from LR face images and restores high-resolution images via efficient feature fusion.
The Facial Attribute Capsule Network~\cite{FACN_AAAI2020} converts the extracted LR face image features into a set of facial attribute capsules by the proposed capsule generation block and utilizes the facial attribute information from both semantic space and probability space to generate the corresponding HR results.
However, since they are trained on synthetic images, these discriminative learning based methods cannot be well generalized to real-world scenarios.

% \begin{figure*}[t!] 
% \flushleft
% \includegraphics[width=\linewidth]{Imgs_FaceSR/tsne_0411.pdf}
% \vspace{-8mm}
% % Visualization
% \caption{\textbf{Distributions of the feature maps extracted by ResNet-101~\cite{resnet_cvpr2016} from HR and SR face images using t-SNE~\cite{TSNE2008}}.
% %
% (a) Visualization of the HR face images.
% %
% (b) Visualization of the SR face images restored by fully-cycled CycleGAN~\cite{cyclegan2017}.
% %
% (c) Visualization of the SR face images restored by our semi-cycled SCGAN.
% %
% Our semi-cycled architecture better retains the feature maps of the SR face images compared to the fully-cycled CycleGAN.
% }
% \label{fig:tsne}
% \end{figure*}

\begin{figure*}[t] 
\centering
	\begin{overpic}[width=\textwidth]{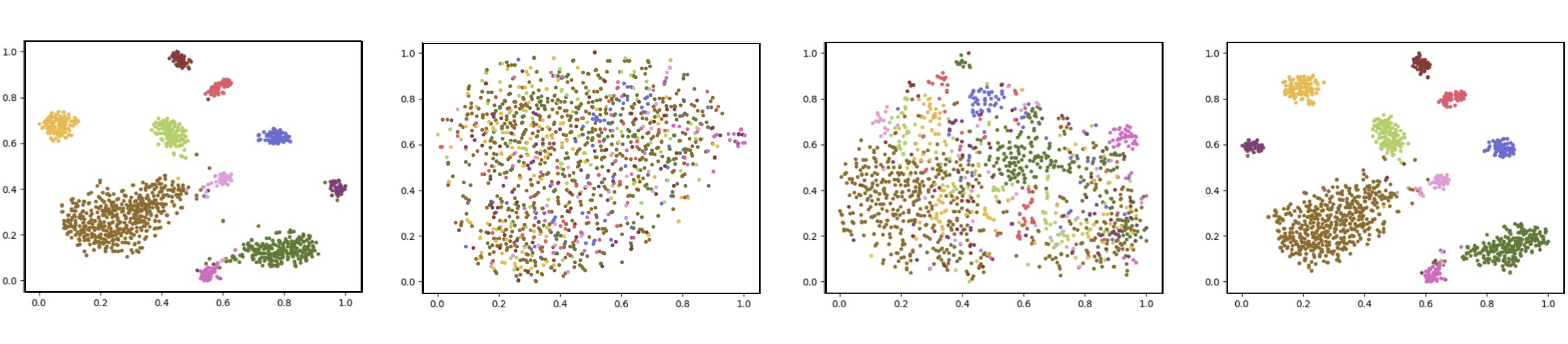}

	\put(7.6,20){\textbf{\scriptsize{HR face images}}}
	\put(11.4,0.4){\textbf{\footnotesize{(a)}}}
 
   	\put(34.4,20){\textbf{\scriptsize{PULSE}~\cite{pulse_cvpr2020}}}
	\put(36.6,0.4){\textbf{\footnotesize{(b)}}}
 
	\put(59,20){\textbf{\scriptsize{Fully-cycled~\cite{cyclegan2017}}}}
	\put(62.4,0.4){\textbf{\footnotesize{(c)}}}
 
   	\put(85.2,20){\textbf{\scriptsize{Semi-cycled}}}
	\put(87.8,0.4){\textbf{\footnotesize{(d)}}}
%   \put(55,10){\includegraphics[width=\textwidth]{Imgs/tsne.pdf}}
    \end{overpic}

\caption{\textbf{Distributions of the feature maps extracted by ResNet-101~\cite{resnet_cvpr2016} from HR and SR face images using t-SNE~\cite{TSNE2008}}.
(a) Visualization of the HR face images.
(b) Visualization of the SR face images restored by state-of-the-art face SR method PULSE~\cite{pulse_cvpr2020}.
(c) Visualization of the SR face images restored by fully-cycled CycleGAN~\cite{cyclegan2017}.
(d) Visualization of the SR face images restored by our semi-cycled SCGAN.
Our semi-cycled architecture better retains the feature maps of the SR face images compared to the fully-cycled CycleGAN.
}
\label{fig:tsne}
\vspace{-3mm}
\end{figure*}
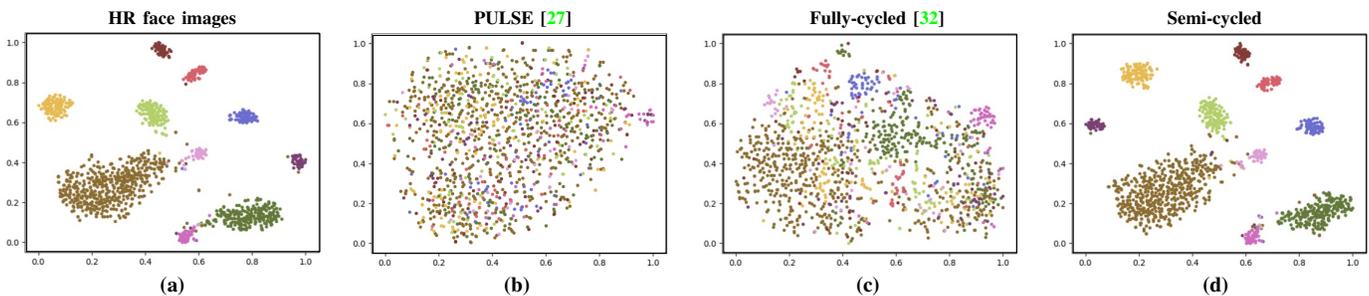

% \begin{figure*}[t!] 
% \flushleft
% \includegraphics[width=\linewidth]{Imgs_FaceSR/FullyandSemi_0420.pdf}
% \vspace{-8mm}
% \caption{\textbf{Comparison of the degradation and restoration results between the fully-cycled CycleGAN architecture~\cite{cyclegan2017} and the proposed semi-cycled architecture of SCGAN.} Where the FID (lower is better) is calculated from the degraded LR face image and the real LR face image, and the PSNR and SSIM are calculated from the restored HR face image and the original HR face image.
% %
% Please zoom in for better view.
% }
% \label{fig:coupleandsemicouple}
% \vspace{-5mm}
% \end{figure*}

Generative models like Generative Adversarial Networks (GANs)~\cite{GAN2014} have achieved remarkable progress on face SR~\cite{LRGAN,maeda2020unpaired,chen2020unsupervised,guo2020closed,zhang2020deblurring}.
URDGN~\cite{URDGN} is among the first work in this direction, but sensitive to the LR face images with large face rotations or poses.
To alleviate this problem, Super-FAN~\cite{superfan_CVPR2018} locates the key points of faces via heat map regression to deal with faces in different angles and poses, which needs large-scale annotations of face landmarks for model training.
LRGAN~\cite{LRGAN} is an unsupervised face SR network by utilizing the architecture of cycle consistency~\cite{cyclegan2017}.
But this method only exploits the consistency within the HR face images while ignoring the consistency within the LR ones.
PULSE~\cite{pulse_cvpr2020} often loses spatial information and identity consistency of face images, by randomly sampling the low-dimensional latent codes. 
The methods of GLEAN~\cite{GLEAN_CVPR2021}, GFPGAN~\cite{GFPGAN2021} and GPEN~\cite{gpen_cvpr2021} utilize a pre-trained StyleGAN~\cite{StyleGAN_cvpr2019} model for face SR, but show limited performance on LR face images with severe degradation.
In this work, we propose to learn three forward or backward mappings, \ie, two independent ``learning-to-degrade'' branches and one shared ``learning-to-SR'' branch, which are semi-cycled to maintain the consistency of both the HR and LR face image reconstructions.
\vspace{-2mm}
\subsection{Generative Adversarial Networks}
Generative Adversarial Networks (GANs)~\cite{GAN2014} have been widely utilized in unsupervised computer vision tasks with great success~\cite{InfoGAN_NIPS2016,cgan2014,pix2pix_cvpr2017,pix2pixhd_cvpr2018,cyclegan2017,dualgan2018,DiscoGAN_ICML2017,LRGAN,maeda2020unpaired,chen2020unsupervised,guo2020closed,zhang2020deblurring}.
InfoGAN~\cite{InfoGAN_NIPS2016} learns explainable feature representation by decomposing the input noise vector into incompressible noise and latent codes, to control semantic features of the generated images.
Conditional GAN (cGAN)~\cite{cgan2014} adds to the original GAN an extra training supervision, achieving great success on image translation tasks~\cite{pix2pix_cvpr2017,pix2pixhd_cvpr2018}.
With the insight of cycle consistency, the methods of CycleGAN~\cite{cyclegan2017}, DualGAN~\cite{dualgan2018}, and DiscoGAN~\cite{DiscoGAN_ICML2017} achieve promising performance on image translation tasks.
This insight has also been resorted by many image restoration methods~\cite{LRGAN,maeda2020unpaired,chen2020unsupervised,guo2020closed,zhang2020deblurring,Drawing_TPAMI2022}.
Among them, LRGAN~\cite{LRGAN} introduces two cycle-consistent generators~\cite{cyclegan2017} for face SR: a ``learning-to-degrade'' branch for HR image degradation and a ``learning-to-SR'' branch for LR face image super-resolution.
However, the two branches are only coupled for HR face image reconstruction, bringing a potential gap between unpaired LR and HR face images.
In this work, we also exploit the powerful generative capability of CycleGAN~\cite{cyclegan2017} for unsupervised real-world face SR.
%  (third)
Built upon LRGAN~\cite{LRGAN}, our SCGAN introduces an additional ``learning-to-degrade'' branch to degrade the super-resolved face images, which are supervised by the real-world LR ones.
\vspace{-3mm}
\subsection{Cycle-Consistent Learning}
The framework of cycle-consistent learning has been developed originally for image-to-image translation~\cite{cyclegan2017} to  jointly learn a paired of coupled branches under the process of backward domain transfer.
From then on, researchers have exploited the cycle-consistent learning framework for many vision tasks such as image restoration~\cite{LRGAN,maeda2020unpaired,chen2020unsupervised,guo2020closed,zhang2020deblurring}.
For example, the methods of~\cite{maeda2020unpaired,chen2020unsupervised} simultaneously perform degradation on the LR images and also restoration on the degraded LR images with pseudo-supervision.

CinCGAN~\cite{CinCGAN_CVPRW2018} first uses an inner CycleGAN to map a noisy LR image into a clean LR image, and then uses an outer CycleGAN to map the clean LR image into an HR one.
MCinCGAN~\cite{MCinCGAN_TIP2019} obtains SR results with different upsampling factors by adjusting the number of recurrent GAN models.
Lugmayr \etal~\cite{lugmayr2019unsupervised} first utilizes the cycle consistency model to degenerate the HR image into a simulated LR image and then forms data pairs for supervised learning of the SR network.
Similar ideas have also been studied in~\cite{sun2021learning,maeda2020unpaired,chen2020unsupervised}.~\cite{kim2020dual} replaces the adversarial loss in CycleGAN~\cite{cyclegan2017} with a dual back-projection loss to form an internal learning framework.

Guo \etal~\cite{guo2020closed} introduced a U-Net like cycle-consistent network for image super-resolution, while Zhang \etal~\cite{zhang2020deblurring} employed the cycle-consistent learning for image deblurring.
Yi \etal~\cite{Drawing_TPAMI2022} proposed an asymmetric cycle-consistent architecture for face portrait line drawing, which is improved by a pre-trained Inception-V3~\cite{Inceptionv3_CVPR2016} under a knowledge distillation scheme~\cite{distilling_nips2015}.
The method of~\cite{kim2020unsupervised} employs the cycle-consistent architecture to firstly convert the input real-world LR image into a pseudo-clean LR image, and then produce the final SR result.
Differently, our SCGAN simultaneously restores the real-world and synthetic LR face images to real-world HR face images.

Cycle-consistent learning is also effective for face SR.
LRGAN~\cite{LRGAN} first learns to degrade the real-world HR face images to the synthetic LR ones by a ``learning-to-degrade’’ sub-network, and then learns to restore the synthetic/real-world LR face images to the corresponding SR ones by a ``learning-to-SR’’ sub-network.
In this paper, we also employ the cycle-consistent learning framework.
% and propose semi-cycled GANs for face SR
But different from fully-cycled CycleGAN, our semi-cycled SCGAN alleviates the adverse gap between real-world HR face images and SR ones by establishing independent degradation mappings.

\section{Proposed Method}
\label{sec:method}
In this section, we introduce the motivation of our Semi-Cycled Generative Adversarial Networks (SCGAN) for unsupervised face super-resolution (SR) in~\S\ref{sec:motivation}.
Then we overview our SCGAN in \S\ref{sec:overview}.
We present three degradation and restoration branches in \S\ref{sec:branch1}, \S\ref{sec:branch2}, and \S\ref{sec:branch3}, respectively.
Finally, the implementation details are provided in \S\ref{sec:implementation}.

\begin{figure*}[t!] 
\flushleft
\includegraphics[width=\linewidth]{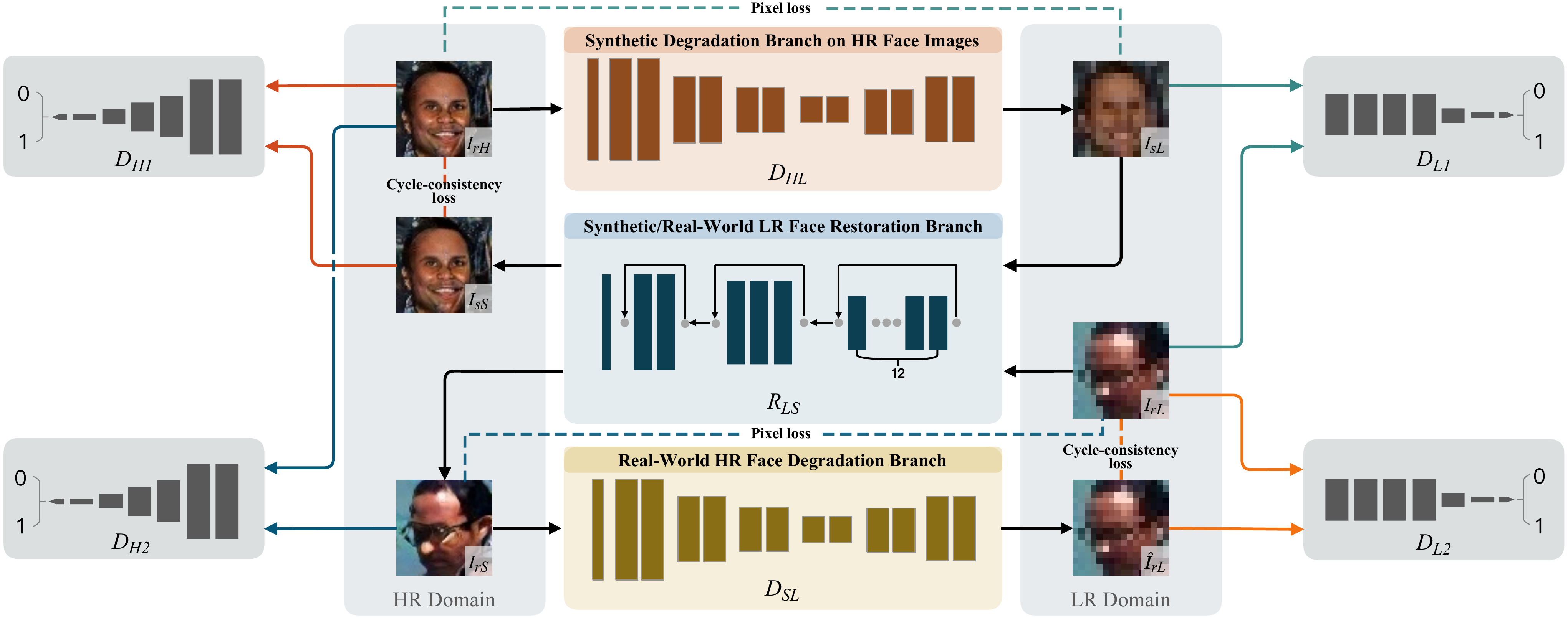}
\vspace{-6mm}
\caption{\textbf{Architecture of our Semi-Cycled Generative Adversarial Network (SCGAN) for unsupervised face super resolution}.
Given a real-world HR face image $\mathbf{I}_{rH}$, we first perform image degradation through the HR face degradation branch $\mathcal{D}_{HL}$ and compute the pixel loss between the downsampled $\mathbf{I}_{rH}$ and the obtained $\mathbf{I}_{sL}$ to preserve more details. Then $\mathbf{I}_{sL}$ is sent to the sub-network $R_{LS}$ to perform SR to obtain $\mathbf{I}_{sS}$. Here, we calculate the cycle consistency loss for $\mathbf{I}_{rH}$ and $\mathbf{I}_{sS}$ to maintain their identity consistency. The above process forms a forward cycle consistent GAN model. Among them, $\mathcal{D}_{L1}$ is responsible for distinguishing $\mathbf{I}_{sL}$ and unpaired input LR face image $\mathbf{I}_{rL}$, $\mathcal{D}_{H1}$ is responsible for distinguishing $\mathbf{I}_{rH}$ and paired SR result $\mathbf{I}_{sS}$. The backward cycle consistent GAN model is similar to the forward one. The most important difference with CycleGAN ~\cite{cyclegan2017} is that, the backward model has an independent degradation sub-network $\mathcal{D}_{SL}$, while not in the fully-cycled situation.}
\label{fig:framework}
\vspace{-3mm}
\end{figure*}

\vspace{-3mm}
\subsection{Motivation}
\label{sec:motivation}

Our goal is to super-resolve real-world low-resolution (LR) face images into the identity preserving high-resolution (HR) face images, without the corresponding paired real-world HR face images.
This task can be suitably tackled under the unsupervised cycle-consistent framework like CycleGAN~\cite{cyclegan2017}.
%  (\ie, $x \rightarrow D(x) \rightarrow x=R(D(x))$)
% (\ie, $y \rightarrow R(y) \rightarrow y=D(R(y))$)
With two fully-cycled generators, CycleGAN well preserves the consistency within the bidirectional translation between two different image domains.
However, the fully-cycled CycleGAN is prone to get stuck upon real-world unsupervised face SR with unpaired LR and HR face images, since the complex degradation in real-world LR face images can hardly be well simulated by the generator simultaneously synthesizing the HR face degradation.
Therefore, directly employing the fully-cycled architecture for real-world face SR inadvertently suffers from an inevitable problem on the degradation gap between synthetic LR images and real-world LR images.
To address this problem, it is natural to model the synthetic and real-world degradations by different generators.
To this end, our SCGAN is developed with two different degradation branches and one restoration branch to learn semi-cycled forward and backward cycle-consistent reconstruction processes.
Our SCGAN is more flexible than the fully-cycled architecture with more accurate unsupervised real-world face SR performance.
Besides, the two independent degradation branches in our SCGAN further facilitate our SCGAN to learn a stronger restoration branch for LR face image, making our SCGAN very robust on super-resolving real-world LR face images.

To globally compare our semi-cycled SCGAN with the fully-cycled CycleGAN, we perform degradation and restoration on HR face images of 10 identities from the LFW dataset~\cite{LFW_2008}, by the same strategy mentioned above.
In Figure~\ref{fig:tsne}, we visualize the input HR face images, and the HR face images restored by PULSE~\cite{pulse_cvpr2020}, CycleGAN~\cite{cyclegan2017} and our SCGAN via t-SNE~\cite{TSNE2008}, using the one-dimensional vectors output by the last fully connected layer of a pre-trained Resnet-101~\cite{resnet_cvpr2016}.
One can see that the distribution of HR face images restored by our SCGAN is more consistent than those restored by PULSE or CycleGAN, with the distribution of input HR face images.
This validates the superiority of our SCGAN over the fully-cycled CycleGAN on the identity preservation of HR face image restoration.

\begin{figure*}[t] 
\centering
	\begin{overpic}[width=\textwidth]{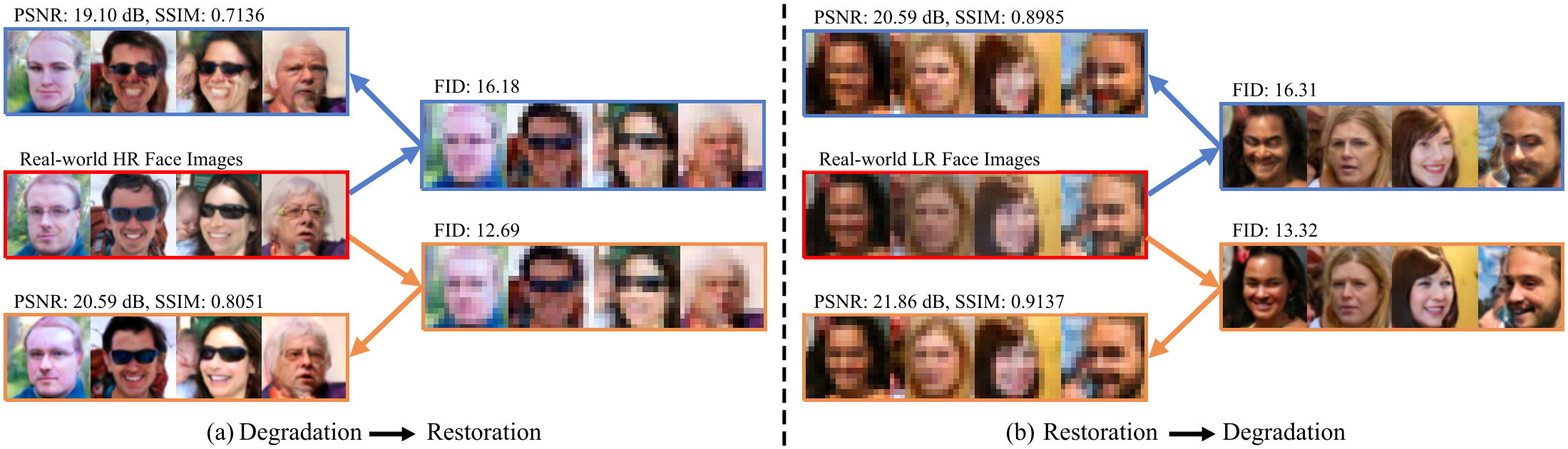}
	\put(24.1,22.4){\textbf{\scriptsize{$\mathcal{R}$}}}
 	\put(24.1,16.2){\textbf{\scriptsize{$\mathcal{D}$}}}
  	\put(23.1,13.2){\textbf{\scriptsize{$\mathcal{D}_{HL}$}}}
   	\put(23.8,6.2){\textbf{\scriptsize{$\mathcal{R}_{LS}$}}}
    
	\put(75.1,22.4){\textbf{\scriptsize{$\mathcal{D}$}}}
 	\put(75.1,16.2){\textbf{\scriptsize{$\mathcal{R}$}}}
  	\put(74.1,13.2){\textbf{\scriptsize{$\mathcal{R}_{LS}$}}}
   	\put(74.8,6.2){\textbf{\scriptsize{$\mathcal{D}_{SL}$}}}
%   \put(55,10){\includegraphics[width=\textwidth]{Imgs/tsne.pdf}}
    \end{overpic}
\vspace{-7mm}
\caption{\textbf{Comparison of the degradation and restoration results between the fully-cycled CycleGAN~\cite{cyclegan2017} and the proposed semi-cycled SCGAN}. The input images are circled with red borders. 
(a) The ``degradation-restoration'' process of the real-world HR face image.
CycleGAN (or Our SCGAN) degrades a real-world HR face image by a degradation branch $\mathcal{D}$ (or $\mathcal{D}_{HL}$) and restores the degraded image by a restoration branch $\mathcal{R}$ (or $\mathcal{R}_{LS}$).
The FID (lower is better) is calculated between the degraded LR face image and the real-world LR face image, the PSNR and SSIM (higher is better) are calculated between the restored HR face image and the original HR face image.
(b) The ``restoration-degradation'' process of the real-world LR face image.
CycleGAN (or Our SCGAN) restores a real-world LR face image by a restoration branch $\mathcal{R}$ (or $\mathcal{R}_{LS}$) and degrades the restored image by a degradation branch $\mathcal{D}$ (or $\mathcal{D}_{SL}$).
The FID is calculated between the restored HR face image and the real HR face image, and the PSNR and SSIM are calculated between the degraded LR face image and the original LR face image.
Please zoom in for better view.}
\vspace{-2mm}
\label{fig:coupleandsemicouple}
\end{figure*}
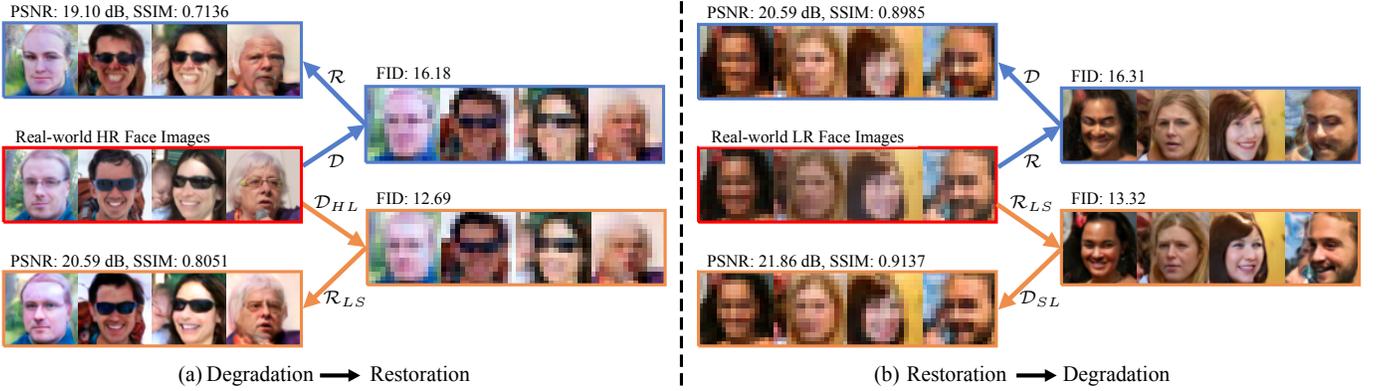

\vspace{-3mm}
\subsection{Network Overview}
\label{sec:overview}
Our SCGAN contains two semi-cycled sub-networks consisting of two independent degradation branches coupled by a restoration branch.
The overall network architecture is illustrated in Figure~\ref{fig:framework}.
The synthetic degradation branch $\mathcal{D}_{HL}$ and the restoration branch $\mathcal{R}_{LS}$ together perform forward cycle-consistent HR face image reconstruction, while the restoration branch $\mathcal{R}_{LS}$ and the real-world degradation branch $\mathcal{D}_{SL}$ together implement the backward cycle-consistent LR face image reconstruction.
The two reconstruction sub-networks are semi-cycled to avoid the adverse effects of the domain gap between the synthetic and realistic LR face images, and to achieve robust yet accurate face SR performance.

\noindent
\textbf{Synthetic HR image degradation branch}.
The HR face image degradation branch, denoted as $\mathcal{D}_{HL}$, degrades an HR face image $\mathbf{I}_{rH}$ to a synthetic LR face image.
% \mathbf{I}_{sL}=
It is the degradation stage of the forward cycle-consistency learning process $\mathbf{I}_{rH}\rightarrow\mathcal{D}_{HL}(\mathbf{I}_{rH})\rightarrow\mathbf{I}_{rH}$, in which the corresponding restoration stage is implemented by the LR face image restoration branch $\mathcal{R}_{LS}$ introduced as follows.

\noindent
\textbf{LR face restoration branch}.
This branch is to enhance the quality of the synthetic LR face image $\mathbf{I}_{sL}$ generated by previous degradation branch $\mathcal{D}_{HL}$ and the real-world LR face image $\mathbf{I}_{rL}$ that is the input in the test stage.
The restoration of synthetic LR face image comprises the forward cycle-consistent learning process ``$\mathbf{I}_{rH}\rightarrow \mathcal{D}_{HL}(\mathbf{I}_{rH})\rightarrow \mathcal{R}_{LS}(\mathcal{D}_{HL}(\mathbf{I}_{rH}))$'', together with the previous synthetic degradation branch, and simultaneously comprises the backward cycle-consistency learning process ``$\mathbf{I}_{rL}\rightarrow \mathcal{R}_{LS}(\mathbf{I}_{rL})\rightarrow \mathcal{D}_{SL}(\mathcal{R}_{LS}(\mathbf{I}_{rL}))$'', in which the corresponding degradation stage is implemented by the real-world HR face image degradation branch $\mathcal{D}_{SL}$ introduced as follows.

\noindent
\textbf{Real-world HR face degradation branch}.
Since the real-world and synthetic LR face images suffer from an inevitable degradation gap, it is reasonable to separately degrade the real-world HR face image $\mathbf{I}_{rH}$ and the synthetic one $\mathbf{I}_{sH}$ generated from the restoration branch $\mathcal{R}_{LH}$ by respective branches.
The real-world HR face degradation branch, with the restoration one, comprise the backward cycle-consistent learning process ``$\mathbf{I}_{rL}\rightarrow \mathcal{R}_{LH}(\mathbf{I}_{rL})\rightarrow \mathcal{D}_{SL}(\mathcal{R}_{LH}(\mathbf{I}_{rL}))$''.

\noindent

\textbf{Discussion}. We train the fully-cycled CycleGAN and our semi-cycled SCGAN with unpaired HR face images from the FFHQ dataset~\cite{StyleGAN_cvpr2019} and LR ones from the Widerface dataset~\cite{widerface_cvpr2016}.
To achieve better reconstruction quality, we train the fully-cycled CycleGAN and our semi-cycled SCGAN with an additional pixel-wise loss function, which will be introduced in~\S\ref{sec:branch1}.
In Figure~\ref{fig:coupleandsemicouple}, we first compare the degraded LR images and the restored HR ones by CycleGAN and our SCGAN, respectively, on four typical real-world HR face images.
The results of FID scores~\cite{FID_2017}, PSNR, and SSIM~\cite{SSIM_TIP2004} are also provided as references.
One can see that the synthetic LR face images degraded from the HR ones in our SCGAN obtains lower FID score than those degraded in the fully-cycled CycleGAN, indicating that our degradation branch obtains more realistic LR face images than those generated by CycleGAN.
To evaluate the reconstruction consistency, we also restore the two sets of synthetic LR face images by the corresponding restoration branches in CycleGAN and our SCGAN, respectively.
Besides, we compare the restored SR images and the degraded LR ones by CycleGAN and our SCGAN, respectively, on four typical real-world LR face images.
We observe that the face images restored by our SCGAN show clear improvement over those restored by CycleGAN on details recovery.
All these results indicate the advantage of our semi-cycled SCGAN over the fully-cycled CycleGAN on unsupervised real-world face SR.

% And the LR face image degraded from $\mathbf{I}_{sHR}$ should be very close to the original LR images.

% In the backward learning process, the SR face degradation branch performs degradation on the restoration results corresponding to the LR face images in the training set from the LR face restoration branch, and generates corresponding degradation results.
\vspace{-3mm}
\subsection{Synthetic Degradation Branch on HR Face Image}
\label{sec:branch1}
This branch, denoted as $\mathcal{D}_{HL}$, aims to learn the degradation process from real-world HR face images to synthetic LR ones.
Given a real-world HR face image $\mathbf{I}_{rH}\in\mathbb{R}^{H\times W\times3}$, we randomly generate a noise vector $\bm{z}\in\mathbb{R}^{HW}$, reshape it into the size of $H\times W$, and concatenate it with $\mathbf{I}_{rH}$ along the channel dimension.
This is to simulate different degrees and types of noise contained in real-world LR face images, as suggested in~\cite{cgan2014}.
The concatenated tensor $[\mathbf{I}_{rH},\bm{z}]\in\mathbb{R}^{H\times W\times4}$ is then fed into the degradation branch $\mathcal{D}_{HL}$ to produce a synthetic LR face image $\mathbf{I}_{sL}$:
\begin{equation}
\mathbf{I}_{sL} = \mathcal{D}_{HL}([\mathbf{I}_{rH},\bm{z}], \bm{\Theta}_{HL}),
\end{equation}
where $\bm{\Theta}_{HL}$ is the set of learnable parameters for $\mathcal{D}_{HL}$.

\begin{figure*}[t]
\flushleft
\includegraphics[width=\linewidth]{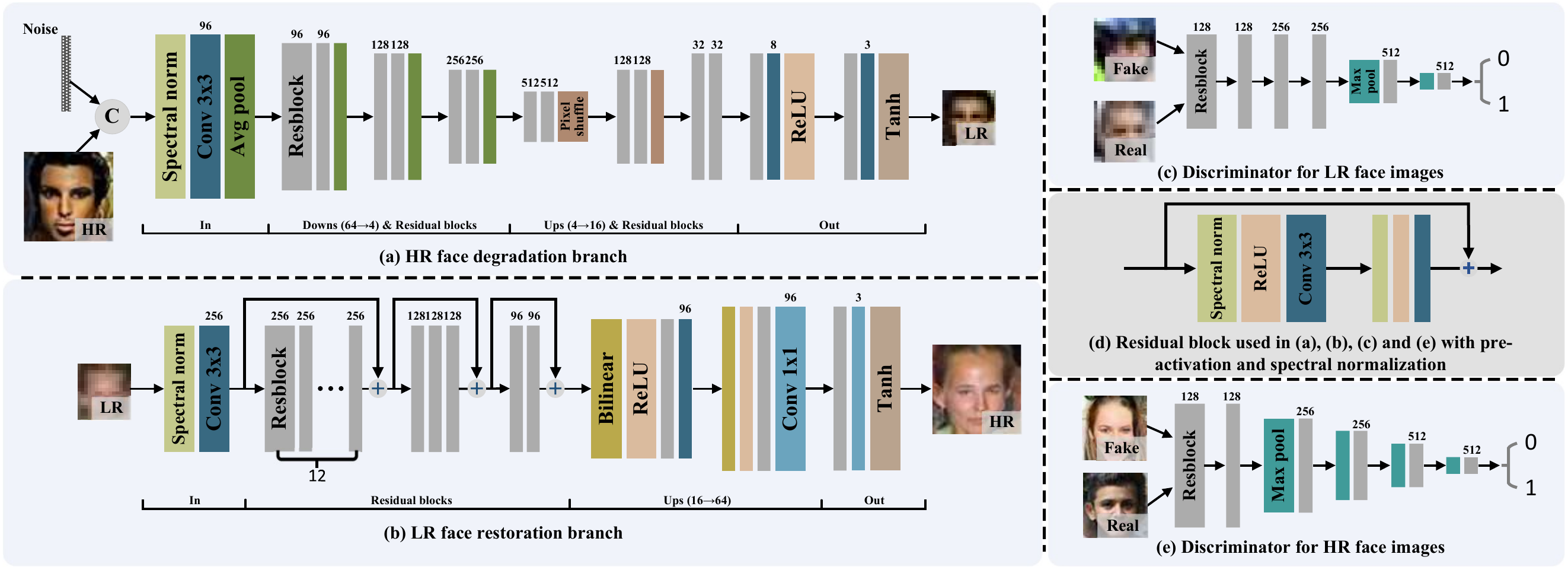}
\vspace{-8mm}
\caption{\textbf{Architectures of the synthetic and real-world HR face degradation branches $\mathcal{D}_{HL}$ and $\mathcal{D}_{SL}$ (a), and the LR face restoration branch $\mathcal{D}_{LS}$ (b)}.
The discriminators are shown in \textbf{(c)} and \textbf{(e)}.
The residual block used in them is shown in \textbf{(d)}.
Please zoom in for the best view.}
\label{fig:netconfig}
\vspace{-5mm}
\end{figure*}

% The structure of our degradation branch $\mathcal{D}_{HL}$ is similar to that in LRGAN~\cite{LRGAN}.
%
As shown in Figure~\ref{fig:netconfig} (a),
the synthetic degradation branch $\mathcal{D}_{HL}$ is in an encoder-decoder architecture.
The encoder begins with a Spectral Normalization (SN)~\cite{SNGAN_ICLR2018}, followed by a $3\times3$ convolutional layer (conv.) and a global average pooling (GAP).
Then six residual blocks (Resblocks) are used to extract meaningful feature.
As shown in Figure~\ref{fig:netconfig} (d), the ResBlock used in $\mathcal{D}_{HL}$ contains two successive sets of SN, ReLU, and $3\times3$ conv., with a skip connection for feature addition.
Here, we use SN to mitigate unstable model training and gradient explosion with the 1-Lipschitz constraint~\cite{SNGAN_ICLR2018}.
GAP is used after every two ResBlocks to reduce feature resolution by a factor of 2.
The decoder also has six Resblocks, with two Pixel-Shuffle operations used after the second and fourth ResBlocks to upsample feature resolution by a factor of 2.
Finally, this branch has two groups of Resblock and $3\times3$ Conv., followed by a ReLU or a Tanh function for nonlinear activation, respectively, and outputs the degraded LR face image $\mathbf{I}_{sL}$.

% instead of Batch Normalization~\cite{BN_ICML2015}

% \textcolor{red}{
% First, the number of $\mathbf{I}_{rH}$ channels is increased to 96 through a $3\times3$ convolutional layer, and the resolution is reduced by an average pooling layer with a scaling factor of 2. After that, it can be divided into 6 groups of residual blocks according to the number of channels, among which, there are 3 groups in the encoder stage, and each group has an average pooling layer with a scaling factor of 2 at the end to reduce the size of the HR face image; In the decoder stage, there are 3 groups of residual blocks, the first two groups contain a pixel-shuffle layer at the end, respectively, to increase the size of the feature map. Finally, the residual block is combined with $3\times3$ convolutional layer to reduce the number of channels to 3, then output the degraded LR face image $\mathbf{I}_{sL}$. Figure~\ref{fig:netconfig} (d) shows the structure of the residual blocks used in $\mathcal{D}_{HL}$.
% }

%motivation
To approximate the degradation in real-world LR face images, the synthetic degradation branch $\mathcal{D}_{HL}$ is learned with an adversarial loss function and a pixel loss function as:
\begin{equation}
    l_{\mathcal{D}_{HL}} = \alpha l_{adv}^{\mathcal{D}_{L1}}
    +
    \beta l_{pix}^{\mathbf{I}_{sL}},
    \label{ld1}
\end{equation}
where $\alpha$ and $\beta$ are the weights of the two loss functions.

\noindent
\textbf{Adversarial loss} $l_{adv}^{\mathcal{D}_{L1}}$ uses
a discriminator $\mathcal{D}_{L1}$ to predict the real-world LR face image $\mathbf{I}_{rL}$ as 1 and the synthetic LR one $\mathbf{I}_{sL}$ as 0, respectively.
As shown in Figure~\ref{fig:netconfig} (c), the discriminator $\mathcal{D}_{L1}$ contains six Resblocks followed by a fully connected layer.
The max-pooling is used before the last two Resblocks to reduce the resolution of the feature map.
% For the adversarial loss we experimented with the recent advancements in the filed (see also~\S\ref{sec:ablation}) and choose the hinge loss~\cite{SNGAN_ICLR2018} for faster training as follows,
% We experimentally compare various existing adversarial losses, and we find that, all of them achieve similar training performance.
Similar to~\cite{SNGAN_ICLR2018}, we use the hinge loss as follows,
\begin{equation}
\begin{split}
    l_{adv}^{\mathcal{D}_{L1}}=\mathbb{E}_{I_{rL}\sim \mathbf{P}_{rL}}[\min(0,\mathcal{D}_{L1}(I_{rL})-1)]+\\
    \mathbb{E}_{I_{sL}\sim \mathbf{P}_{sL}}[\min(0,-1-\mathcal{D}_{L1}(I_{sL}))],
\end{split}
\label{adversariallossh2l}
\end{equation}
where $\mathbf{P}_{rL}$ and $\mathbf{P}_{sL}$ are the distributions of real-world LR face image $\mathbf{I}_{rL}$ and the synthetic one $\mathbf{I}_{sL}$ degraded by $D_{HL}$ from the real-world HR face image $\mathbf{I}_{rH}$, respectively.

\noindent
\textbf{Pixel loss} $l_{pix}^{\mathbf{I}_{sL}}$ is calculated between the synthetic degradation image $\mathbf{I}_{sL}$ and the input HR face image $\mathbf{I}_{rH}$ downsampled to the same resolution with $\mathbf{I}_{sL}$ by average pooling.
Here, we adopt the $\ell_1$ loss function that is widely used in image SR task~\cite{LRGAN,tmnet} to well recover image details.
%\begin{equation}
%    l_{pix} = \frac{1}{WH}\sum_{i=1}^{W}\sum_{j=1}^{H}\left |(\mathbf{I}_{rH})^{i, j} - \mathbf{I}_{LR})^{i,j}\right|,
%\label{contentloss}
%\end{equation}
%where $W$, $H$ denote the size of the generated output image, respectively.
% Specifically, for degradation branch $\mathcal{D}_{HL}$, the function $F_{1}$ is implemented using an average pooling layer, while the function $F_{2}$ is just a simple identity function.

\vspace{-3mm}
\subsection{Synthetic/Real-World LR Face Restoration Branch}
\label{sec:branch2}
The LR face restoration branch $\mathcal{R}_{LS}$ is a hub shared by the forward and backward cycle-consistency learning processes.
In the forward learning process, it restores the synthetic LR image $\mathbf{I}_{sL}$ degraded from the HR face image $\mathbf{I}_{rH}$ via $\mathcal{D}_{HL}$, while in the backward learning process, it restores the real-world LR face image $\mathbf{I}_{rL}$.
Denote the SR image restored from $\mathbf{I}_{sL}$ as $\mathbf{I}_{sS}$ and the SR image restored from $\mathbf{I}_{rL}$ as $\mathbf{I}_{rS}$, the restoration process is as follows:
\begin{equation}
\mathbf{I}_{sS} = \mathcal{R}_{LS}(\mathbf{I}_{sL},\bm{\Theta}_{LS}),
\end{equation}
\begin{equation}
\mathbf{I}_{rS} = \mathcal{R}_{LS}(\mathbf{I}_{rL},\bm{\Theta}_{LS}),
\end{equation}
where $\bm{\Theta}_{LS}$ is the learnable parameters of the branch $\mathcal{R}_{LS}$.

As shown in Figure~\ref{fig:netconfig} (b), our restoration branch $\mathcal{R}_{LS}$ also begins with a
Spectral Normalization~\cite{SNGAN_ICLR2018}, followed by a $3\times3$ convolutional layers.
% consists of 2 $3\times3$ convolutional layers, 2 $1\times1$ convolutional layers and 20 residual blocks. For the input $\mathbf{I}_{sL}$/$\mathbf{I}_{rL}$
Then three groups of 12, 3, and 2 Resblocks are used to extract meaningful features, and in each group the input and output of each group have a skip connection for feature addition to preserve high-frequency details.
To enhance its resolution, the feature map is upsampled by a factor of 4 by two bilinear interpolations, followed by a group of ``ReLU-Resblock-$3\times3$ Conv.'' and a group of "ReLU-Resblock-$1\times1$ Conv.", respectively.
Finally, this branch outputs the restored HR face image through a Resblock, a $1\times1$ Conv., and a Tanh activation function.

% Then the feature maps goes through one block of followed by 1 sequence of "residual block-$3\times3$ convolution" and 2 sequence of "residual block-$1\times1$ convolution" operation to output the final restoration result $\mathbf{I}_{sS}$/$\mathbf{I}_{rS}$. 
% 
% During the whole process, the resolution of the feature map is increased 2 times by bilinear Interpolation with a scaling factor of 2.
% 
% The architecture of $\mathcal{R}_{LS}$ is shown in Figure~\ref{fig:netconfig} (b).
% \par
% The architecture of $\mathcal{R}_{LS}$ is shown in Figure~\ref{fig:netconfig} (b), the number of channels of the input LR face image ($\mathbf{I}_{sL}$/$\mathbf{I}_{rL}$) is first increased to 256 by a $3\times3$ convolutional layer, after that, there are 3 groups of residual blocks divided by the number of channels, and the number of residual blocks included is 12, 3, and 2, respectively. Here, to minimize the loss of high-frequency details, each group has a skip connection to connect the input of the first block and the output of the last block. After that, the resolution of the feature map is increased by 2 bilinear upsampling layers with a scaling factor of 2, and the number of channels is reduced to 3 by $3\times3$ and $1\times1$ convolutional layer. Finally, output the restoration result ($\mathbf{I}_{sS}$/$\mathbf{I}_{rS}$). Figure~\ref{fig:netconfig} (d) shows the structure of the residual blocks used in $\mathcal{R}_{LS}$. 

The restoration branch $\mathcal{R}_{LS}$ aims to generate high-quality face images, shared by the forward and backward learning processes.
% Here, we use two sets of loss functions, denoted by $l_{\mathcal{R}_{LS}}^{\mathbf{I}_{sS}}$ and $l_{\mathcal{R}_{LS}}^{\mathbf{I}_{rS}}$.
% $l_{\mathcal{R}_{LS}}^{\mathbf{I}_{rS}}$
We use the combination of adversarial loss $l_{adv}^{\mathcal{D}_{H1}}$ and cycle-consistency loss $l_{cyc}^{\mathbf{I}_{sS}}$ in the forward learning process, and use the combination of adversarial loss  $l_{adv}^{\mathcal{D}_{H2}}$ and pixel loss $l_{pix}^{\mathbf{I}_{rS}}$ in the backward learning process.
% The purpose of the adversarial loss is to drive $\mathcal{R}_{LS}$ to generate realistic restoration result, the pixel loss is to enrich the contextual details of the SR results, the cycle-consistency loss is to helps $\mathcal{R}_{LS}$ maintains the identity information.
% 
The overall loss function for this branch is
\vspace{-1mm}
\begin{equation}
    l_{\mathcal{R}_{LS}} = \theta l_{\mathcal{R}_{LS}}^{\mathbf{I}_{sS}} + \gamma l_{\mathcal{R}_{LS}}^{\mathbf{I}_{rS}},
    \label{lr}
\end{equation}
where $\theta$ and $\gamma$ are the corresponding weights, and
\begin{equation}
    l_{\mathcal{R}_{LS}}^{\mathbf{I}_{sS}}  = \alpha l_{adv}^{\mathcal{D}_{H1}} + \beta l_{cyc}^{\mathbf{I}_{sS}},
\end{equation}
\vspace{-4mm}
\begin{equation}
    l_{\mathcal{R}_{LS}}^{\mathbf{I}_{rS}} = \alpha l_{adv}^{\mathcal{D}_{H2}} + \beta l_{pix}^{\mathbf{I}_{rS}}
    \label{lr2}.
\end{equation}

%The adversarial loss $l_{adv}^{\mathcal{D}_{H1}}$ and $l_{adv}^{\mathcal{D}_{H2}}$, cycle-consistency loss $l_{cyc}^{\mathbf{I}_{sS}}$ and pixel loss $l_{pix}^{\mathbf{I}_{rS}}$ used in this branch are given in detail as follows,
\par
\noindent
\textbf{Adversarial loss}.
We use a discriminator $\mathcal{D}_{H1}$ to predict the real-world HR face image $I_{rH}$ as 1 and the synthetic
SR image $I_{sS}$ as 0, respectively.
Similarly, we use a discriminator $\mathcal{D}_{H2}$ to predict the real-world HR face image $I_{rH}$ as 1 and the real-world SR image $I_{rS}$ as 0, respectively.
The adversarial losses $l_{adv}^{\mathcal{D}_{H1}}$ and $l_{adv}^{\mathcal{D}_{H2}}$ are computed as follows,
\begin{equation}
\begin{split}
    l_{adv}^{\mathcal{D}_{H1}}=\mathbb{E}_{I_{rH}\sim \mathbf{P}_{rH}}[\min(0,\mathcal{D}_{H1}(I_{rH})-1)]+\\
    \mathbb{E}_{I_{sS}\sim\mathbf{P}_{sS}}[\min(0,-1-\mathcal{D}_{H1}(I_{sS})],
\end{split}
\label{adversarialloss_l2h1}
\end{equation}
\begin{equation}
\begin{split}
    l_{adv}^{\mathcal{D}_{H2}}=\mathbb{E}_{I_{rH}\sim \mathbf{P}_{rH}}[\min(0,\mathcal{D}_{H2}(I_{rH})-1)]+\\
    \mathbb{E}_{I_{rS}\sim\mathbf{P}_{rS}}[\min(0,-1-\mathcal{D}_{H2}(I_{rS})].
\end{split}
\label{adversarialloss_l2h2}
\end{equation}
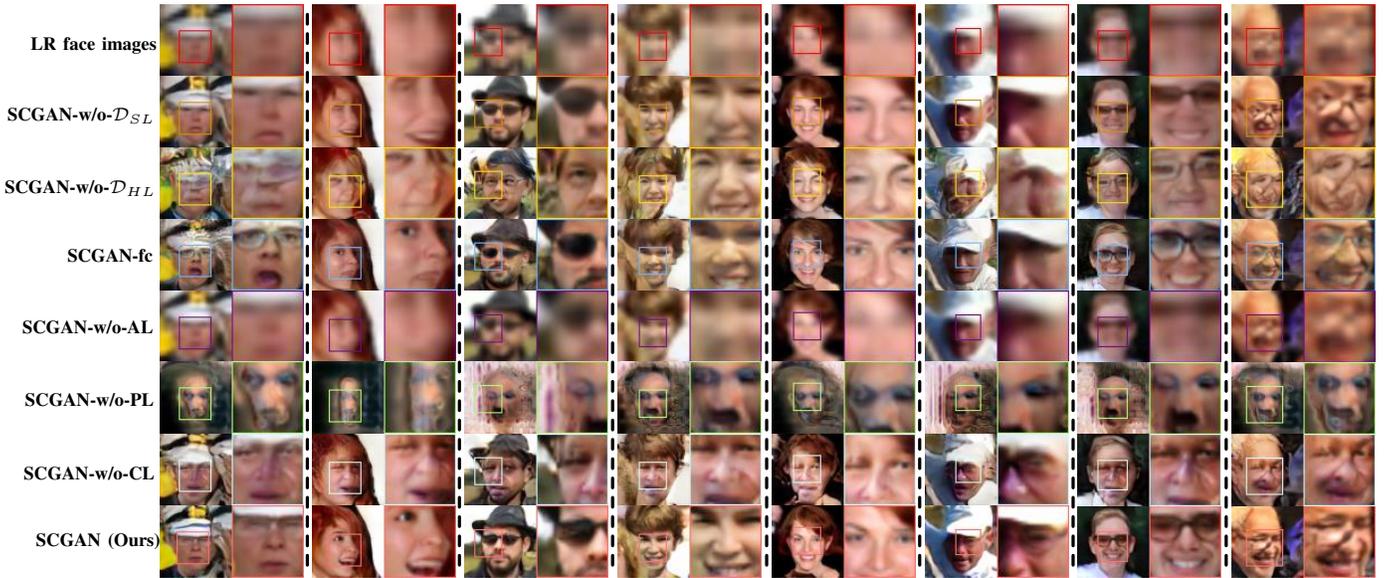
\begin{figure*}[t!] 
\centering
	\begin{overpic}[width=\textwidth]{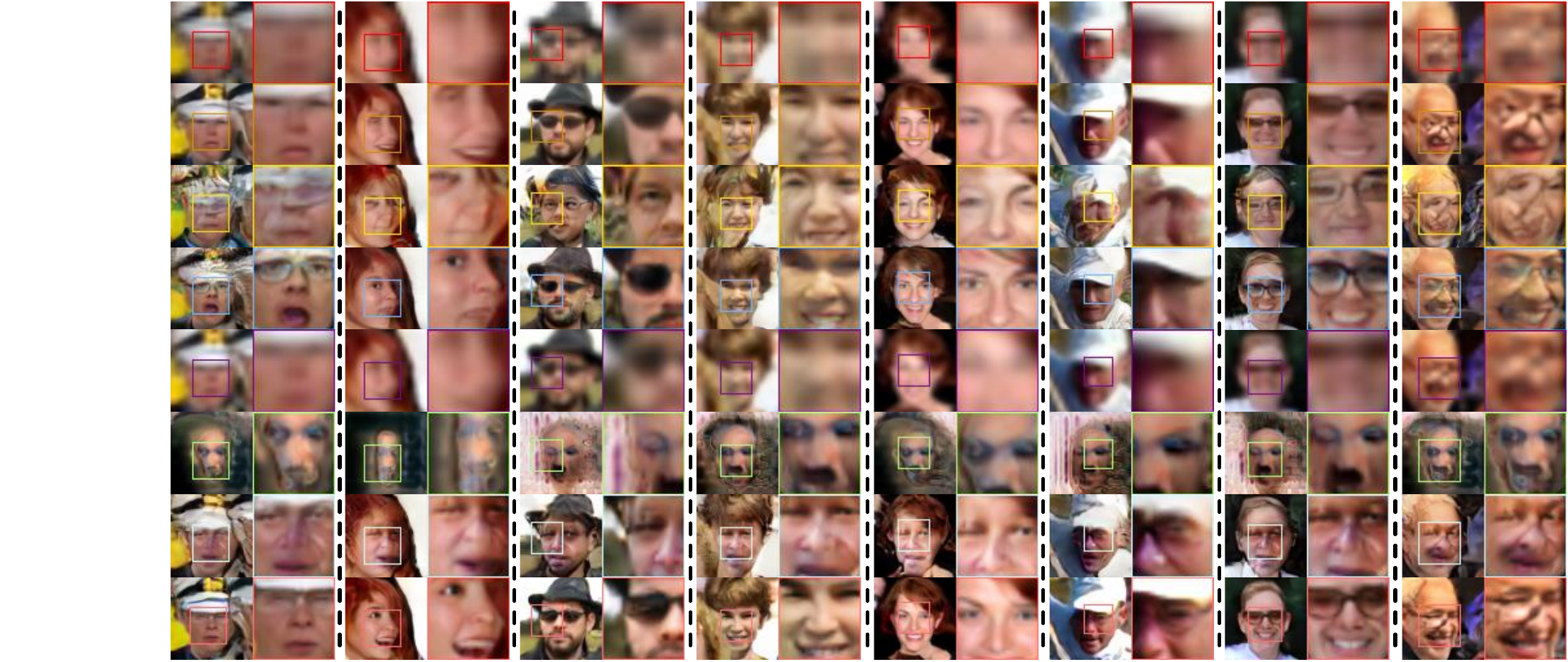}
	\put(1.4,38.8){\textbf{\scriptsize{LR face images}}}
	\put(-0.3,33.6){\textbf{\scriptsize{SCGAN-w/o-$\mathcal{D}_{SL}$}}}
	\put(-0.5,28.3){\textbf{\scriptsize{SCGAN-w/o-$\mathcal{D}_{HL}$}}}
	\put(4.1,23.3){\textbf{\scriptsize{SCGAN-fc}}}
	\put(0.8,18){\textbf{\scriptsize{SCGAN-w/o-AL}}}
	\put(1.0,12.7){\textbf{\scriptsize{SCGAN-w/o-PL}}}
	\put(0.9,7.3){\textbf{\scriptsize{SCGAN-w/o-CL}}}
    \put(1.8,2.4){\textbf{\scriptsize{SCGAN (Ours)}}}
%   \put(55,10){\includegraphics[width=\textwidth]{Imgs_FaceSR/tsne.pdf}}
    \end{overpic}
\vspace{-8mm}
\caption{\textbf{Comparison results by different variants of our SCGAN on representative LR face images from the Widerface~\cite{widerface_cvpr2016} dataset.}}
\label{fig:variants}
\vspace{-5mm}
\end{figure*}

Here, $\mathbf{P}_{rH}$, $\mathbf{P}_{sS}$, and $\mathbf{P}_{rS}$ are the distributions of real-world HR face image $\mathbf{I}_{rH}$, synthetic SR image $\mathbf{I}_{sS}$ restored by $\mathcal{R}_{LS}$ from the synthetic LR face image $\mathbf{I}_{sL}$, and real-world SR image $\mathbf{I}_{rS}$ restored by $\mathcal{R}_{LS}$ from the real-world LR face image $\mathbf{I}_{rL}$, respectively.
The discriminators $\mathcal{D}_{H1}$ and $\mathcal{D}_{H2}$ are in the same structure, which contains six Resblocks followed by a fully connected layer and uses max-pooling before the last four Resblocks, as shown in Figure~\ref{fig:netconfig} (e).

\noindent
\textbf{Cycle-consistency loss} $l_{cyc}^{\mathbf{I}_{sS}}$ is an $\ell_{1}$ loss function used here to make our restoration branch $\mathcal{R}_{LS}$ well preserve the identity information and well recover the face details.
% \begin{equation}
%     l_{cyc} = \frac{1}{WH}\sum_{i=1}^{W}\sum_{j=1}^{H}\left |(\mathbf{I}_{HR})^{i, j} - (\mathbf{I}_{SR})^{i,j}\right |
%     \label{cycleconsistencyloss}
% \end{equation}

\noindent
\textbf{Pixel loss} $l_{pix}^{\mathbf{I}_{rS}}$ is an $\ell_{1}$ loss function to penalize the difference between real-world HR face image $\mathbf{I}_{rH}$ and SR one $\mathbf{I}_{rL}$ (upsampled to the same size of $\mathbf{I}_{rH}$ by bicubic interpolation).

\subsection{Real-World HR Face Degradation Branch}
\label{sec:branch3}

This branch, denoted as $\mathcal{D}_{SL}$, learns to degrade the real-world SR face image $\mathbf{I}_{rS}$ restored from the real-world LR image $\mathbf{I}_{rL}$ via $\mathcal{R}_{SL}$ as follows,
\begin{equation}
\hat{\mathbf{I}}_{rL} = \mathcal{D}_{SL}(\mathbf{I}_{rS},\bm{\Theta}_{SL}),
\end{equation}
where $\bm{\Theta}_{SL}$ is the learnable parameters.
As shown in Figure~\ref{fig:netconfig} (b), the architecture of $\mathcal{D}_{SL}$ is the same as that of the synthetic HR face degradation branch $\mathcal{D}_{HL}$ introduced in \S\ref{sec:branch1}.
% The key difference is, $\mathcal{D}_{SL}$ is responsible for the degradation on the HR face images restored from the real-world LR face images in the backward cycle-consistent reconstruction process, instead of the real-world HR face images.

To make the branch $\mathcal{D}_{SL}$ generate degradation results that are close to real-world LR face images, here, we employ the adversarial loss $l_{adv}^{\mathcal{D}_{L2}}$ and the cycle-consistency loss $l_{cyc}^{\hat{I}_{rL}}$ between the output LR image  $\hat{\mathbf{I}}_{rL}$ and the real-world one $\mathbf{I}_{rL}$, which are computed as follows,
\begin{equation}
    l_{\mathcal{D}_{SL}} = \alpha l_{adv}^{\mathcal{D}_{L2}} + \beta l_{cyc}^{\hat{I}_{rL}}.
\end{equation}

\par
\noindent
\textbf{Adversarial loss}
$l_{adv}^{\mathcal{D}_{L2}}$ uses a discriminator $D_{L_{2}}$ to predict the real-world LR face image $\mathbf{I}_{rL}$ as 1 and the output LR one $\hat{\mathbf{I}}_{rL}$ as 0, respectively.
The architecture of $D_{L2}$ is the same as that of $D_{L1}$ introduced in \S\ref{sec:branch1}.
Similar to Eq.~(\ref{adversariallossh2l}), the adversarial loss $l_{adv}^{\mathcal{D}_{L2}}$ is computed as follows,
\begin{equation}
\begin{split}
    l_{adv}^{\mathcal{D}_{L2}}=\mathbb{E}_{I_{rL}\sim \mathbf{P}_{rL}}[\min(0,\mathcal{D}_{L2}(I_{rH})-1)]+\\ \mathbb{E}_{\hat{I}_{rL}\sim\mathbf{P}_{\hat{rL}}}[\min(0,-1-\mathcal{D}_{L2}(\hat{I}_{rL})],
\end{split}
\label{adversarialloss_h2l2}
\end{equation}
where $\mathbf{P}_{rL}$ and $\mathbf{P}_{\hat{rL}}$ are the distributions of real-world LR face image $\mathbf{I}_{rL}$ and synthetic one $\hat{I}_{rL}$ degraded by $D_{SL}$ from the real-world SR face image $\mathbf{I}_{rS}$, respectively.

\noindent
\textbf{Cycle-consistency loss}
$l_{cyc}^{\hat{I}_{rL}}$ is an $\ell_1$ loss function to penalize the difference between the LR image $\hat{\mathbf{I}}_{rL}$ degraded by this branch and the corresponding real-world LR face image $\mathbf{I}_{rL}$.

% 
% $l_{cyc}^{\hat{I}_{rL}}$ is used to make our degradation branch $D_{SL}$ to maintain the identity information better. Here, we compute the cycle-consistency loss between $\hat{I}_{rL}$ and $\mathbf{I}_{rL}$ in Eq.~\ref{cycleconsistencyloss}.
% 
% Similar to the real-world HR face degradation branch $\mathcal{D}_{HL}$, we feed $\hat{I}_{rL}$ and $\mathbf{I}_{rL}$ together to the discriminator $\mathcal{D}_{L2}$ and compute the adversarial loss to drive $\mathcal{D}_{SL}$ to generate more realistic degradation result.

% Finally, the entire backward learning process consists of $R$ and $\mathcal{D}_{HL}$ connections, so we also calculate the cycle consistency loss between ${I}'_{LR_{2}}$ and $\mathbf{I}_{LR}$ to drive $\mathcal{D}_{HL}$ to ensure content consistency. 

\subsection{Implementation Details}
\label{sec:implementation}
The parameters of all three branches in our SCGAN are initialized by Kaiming initialization~\cite{kaiminginnitial}, and optimized by Adam~\cite{Adam_2014} with $\beta_{1}=0.9$ and $\beta_{2}=0.999$.
We set $\alpha=1,\beta=0.05$ in Eqs.~(\ref{ld1})-(\ref{lr2}) and $\theta=1,\gamma=0.05$ in Eq.~(\ref{lr}).
Our SCGAN is trained for 200 epochs.
The learning rate is initialized as $1\times10^{-4}$ and decayed to $1\times10^{-5}$ with the cosine annealing scheme at every 10 epochs.
The batch size is set as 64.
We implement our SCGAN in PyTorch~\cite{pytorch_nips2019} and train it on a Tesla V100 GPU, which takes about 42 hours.

\vspace{-3mm}

\section{Experiments}
\label{sec:experiments}
In this section, we first introduce the experimental setup, including the dataset and evaluation metrics in \S\ref{sec:dataset}.
We then conduct a comprehensive ablation study in \S\ref{sec:ablation} to validate the role of each component of our SCGAN on face SR.
Comparison with the state-of-the-art methods on real-world face SR are presented in \S\ref{sec:comparisonwithsota}.
%
% In \S\ref{sec:CBCT}, we employ our SCGAN on dental CBCT image super resolution task to validate its effectiveness beyond face SR.
%
Finally, we apply our SCGAN into three other vision tasks, \eg, face detection, face verification, and face landmark detection in \S\ref{sec:application}.
\vspace{-3mm}
\subsection{Dataset and Evaluation Metric} 
\label{sec:dataset}
% To more objectively evaluate the performance of our SCGAN and its ability to real-world face super-resolution, firstly, 
%
\noindent
\textbf{Training set}.
We train our SCGAN, its variants (to be introduced in \S\ref{sec:ablation}), and all the comparison methods (to be introduced in \S\ref{sec:comparisonwithsota}) with the 20,000 high-quality, high-resolution (HR) face images from the real-world FFHQ dataset~\cite{StyleGAN_cvpr2019} and the 4,000 low-quality, low-resolution (LR) face images from the real-world Widerface dataset~\cite{widerface_cvpr2016}.
% Note that the HR and LR images are unpaired.
% For data augmentation, we perform the following random degradation model on real-world LR face images as
%\begin{equation}
%I^{deg}=((I^{L}\otimes k) + n_{\delta })_{JPEG_{q}} 
%    \label{eq:degradation}
%\end{equation}
%For the input LR face image $I^{L}$, it is first convolved with Gaussian blur kernel $k$.
% After that, we add additive white Gaussian noise $n_{\delta}$ to the face image and finally compress it with JPEG with quality factor $q$. For each degradation process, we randomly sample $k$, $\delta$ and $q$ from $\left \{ 0.5:8 \right \}$, $\left \{ 1:25 \right \}$ and $\left \{ 30:95 \right \}$, respectively.
% In addition, inspired by~\cite{BSRGAN_ICCV2021}, we also add camera noise to further improve the degradation diversity of real-world LR face images.

\noindent
\textbf{Test set}.
We evaluate the comparison methods on four popular face SR datasets, including two synthetic datasets, \ie, \emph{LS3D-W balanced}~\cite{LS3D_2017} and \emph{FFHQ}~\cite{StyleGAN_cvpr2019}, and two real-world datasets, \ie, \emph{Widerface}~\cite{widerface_cvpr2016} and our newly collected \emph{Webface}:
% obtained by simple bilinear downsampling and random degradation process in Eq.~\ref{eq:degradation},
\begin{itemize}
    \item \emph{LS3D-W balanced}~\cite{LS3D_2017} contains 7,200 HR face images taken in different scenes and poses.
    We randomly select 1000 face images and perform simple bilinear downsampling to produce synthetic LR face images.
    \item \emph{FFHQ}~\cite{StyleGAN_cvpr2019} contains 70,000 HR face images, 20,000 of which are used as the training set.
    We randomly select 2,500 images from the remaining images to perform random degradation  $I_{sL}=((I_{rL}\otimes k)\downarrow + n_{\delta })_{JPEG_{q}}$ to produce the synthetic LR face images, as suggested in~\cite{LRGAN}.
    Here, $k$ is a Gaussian blur kernel, $\downarrow$ is a downsampling operation randomly chosen from bilinear or bicubic at a scaling factor of 4, $n_{\delta}$ is additive white Gaussian noise, and $JPEG_{q}$ is the JPEG compression with quality factor $q$.
    For each degradation, we randomly sample $k\in[0.5,8]$, $\delta\in[1,25]$, and $q\in[30,95]$, respectively.

    \item \emph{Widerface}~\cite{widerface_cvpr2016} contains 32,203 real-world LR face images from 62 versatile scenes, and we randomly select 2,000 images with unknown yet complex degradation process.
    \item \emph{WebFace}.
    We crawled 1,028 real-world LR face images, with different genders, ages, races, expression, postures, and unknown degradation process, from the internet.
    \item \emph{DroneSURF}~\cite{dronesurf_2019} contains more than 720,000 images with faces from drone-captured videos in the wild, we randomly selected 1,000 images, and directly cropped the patches of size $16\times16$ that contain human faces.
\end{itemize}

\noindent
\textbf{Evaluation metrics}.
We employ feature-level and image-level metrics to objectively and comprehensively evaluate results of different methods.
On all test sets, we use the Frechet Inception Distance (FID)~\cite{FID_2017} and Kernel Inception Distance (KID)~\cite{KID_ICLR2018} to evaluate the distribution distance between the output SR images and real-world HR face images on diversity and visual quality, respectively.
On two synthetic test sets, we also use the Learned Perceptual Image Patch Similarity (LPIPS)~\cite{LPIPS2018} to measure the distance of human perception between the SR images and the corresponding ground-truth ones.
On two real-world test sets, we also use the widely used Natural Image Quality Evaluator (NIQE)~\cite{NIQE_2012} to evaluate the naturalness of restored face images.
Besides, we compute the detection accuracy of the method based on RetinaFace~\cite{retinaface_CVPR2020} on the SR face images from each dataset, which indirectly measures the capability of face SR methods on identity preservation.
\vspace{-3mm}
\subsection{Ablation Study}
\label{sec:ablation}
To study the role of each component in our SCGAN to its effectiveness on real-world face SR, here, we conduct detailed examinations of our SCGAN on different LR face image test sets.
Specifically, we access
a) the benefits of our semi-cycle architecture;
b) whether to share parameters of two degradation branches or not in our SCGAN;
c) how different loss functions (adversarial loss, pixel loss, and cycle consistency loss) contribute to our SCGAN;
d) how different combinations of adversarial losses influence our SCGAN;
e) how different structures of the degradation branch influence our SCGAN; f) how about using two independent restoration branches with a shared degradation branch; g) how to determine the weights of different loss functions.

\begin{table}[t]
\centering
  \footnotesize
  \renewcommand{\arraystretch}{1.2}
  \setlength\tabcolsep{3pt}
  \caption{\small \textbf{Quantitative results on two synthetic and two real-world datasets by our SCGAN and its variants with different architectures}.
  The best, second best, and third best results are highlighted in \textbf{\textcolor{red}{red}}, \textbf{\textcolor{blue}{blue}} and \textbf{bold}, respectively.}
  \vspace{-3mm}
\begin{tabular}{c||l||c|c|c}
\Xhline{1.6pt}
Dataset & Variant & FID $\downarrow$ & KID $\downarrow$ & LPIPS $\downarrow$ \\ \hline\hline
\multirow{4}{*}{\begin{tabular}[c]{@{}c@{}}LS3D-W\\ balanced\\~\cite{LS3D_2017}\end{tabular}}
        & SCGAN-w/o-$\mathcal{D}_{SL}$    
        & 43.24 & 3.64$\pm$0.10 & \textbf{\textcolor{blue}{0.081}}  \\
        & SCGAN-w/o-$\mathcal{D}_{HL}$  
        & \textbf{27.49} & \textbf{3.51$\pm$0.10} & 0.129 \\
        & SCGAN-fc    
        & \textbf{\textcolor{blue}{25.84}} & \textbf{\textcolor{blue}{1.99$\pm$0.07}} & \textbf{0.088}  \\
        & SCGAN 
        & \textbf{\textcolor{red}{22.55}} & \textbf{\textcolor{red}{1.26$\pm$0.06}} & \textbf{\textcolor{red}{0.068}}  \\ \hline\hline
\multirow{4}{*}{\begin{tabular}[c]{@{}c@{}}FFHQ\\ \cite{StyleGAN_cvpr2019}\end{tabular}}
        & SCGAN-w/o-$\mathcal{D}_{SL}$  & 27.57 & \textbf{3.18$\pm$0.09} & \textbf{\textcolor{blue}{0.242}}  \\
        & SCGAN-w/o-$\mathcal{D}_{HL}$  & \textbf{19.25} & 3.44$\pm$0.12 & 0.310  \\
        & SCGAN-fc    & \textbf{\textcolor{blue}{15.15}} & \textbf{\textcolor{blue}{2.00$\pm$0.08}} & \textbf{0.247}  \\
        & SCGAN       & \textbf{\textcolor{red}{9.06}} & \textbf{\textcolor{red}{0.94$\pm$0.05}} & \textbf{\textcolor{red}{0.197}} \\ \hline\hline
Dataset & Variant & FID $\downarrow$ & KID $\downarrow$ & NIQE $\downarrow$ \\ \hline\hline
\multirow{4}{*}{\begin{tabular}[c]{@{}c@{}}Widerface\\ \cite{widerface_cvpr2016}\end{tabular}}
        & SCGAN-w/o-$\mathcal{D}_{SL}$    & 33.96 & \textbf{3.14$\pm$0.11} & \textbf{\textcolor{red}{6.5738}}  \\
        & SCGAN-w/o-$\mathcal{D}_{HL}$  & \textbf{20.53} & 3.36$\pm$0.12 & 6.7653 \\
        & SCGAN-fc    & \textbf{\textcolor{blue}{16.02}} & \textbf{\textcolor{blue}{1.65$\pm$0.06}} & \textbf{6.7538} \\
        & SCGAN       & \textbf{\textcolor{red}{13.32}} & \textbf{\textcolor{red}{1.08$\pm$0.05}} & \textbf{\textcolor{blue}{6.6192}}  \\ \hline\hline
\multirow{4}{*}{\begin{tabular}[c]{@{}c@{}}WebFace \end{tabular}}
        & SCGAN-w/o-$\mathcal{D}_{SL}$    & 40.57 & 3.75$\pm$0.10 & \textbf{\textcolor{red}{6.5697}}  \\
        & SCGAN-w/o-$\mathcal{D}_{HL}$  & \textbf{25.29} & \textbf{3.32$\pm$0.10} & \textbf{6.7358}  \\
        & SCGAN-fc    & \textbf{\textcolor{blue}{23.31}} & \textbf{\textcolor{blue}{2.10$\pm$0.07}} & 6.7464 \\
        & SCGAN       & \textbf{\textcolor{red}{21.06}} & \textbf{\textcolor{red}{1.39$\pm$0.06}} & \textbf{\textcolor{blue}{6.5835}} \\ \hline
\end{tabular}
\label{tab:variants}
\vspace{-7mm}
\end{table}

\begin{table}[t]
\centering
  \footnotesize
  \renewcommand{\arraystretch}{1.2}
  \setlength\tabcolsep{3pt}
  \caption{\small  \textbf{Quantitative results on two synthetic and two real-world datasets by our SCGAN and its variants with different parameter-sharing schemes in two degradation branches}.
  The best, second best, and third best results are highlighted in \textbf{\textcolor{red}{red}}, \textbf{\textcolor{blue}{blue}} and \textbf{bold}, respectively.}
  \vspace{-3mm}
\begin{tabular}{c||c||c|c|c}
\Xhline{1.6pt}
Dataset & Variant & FID $\downarrow$ & KID $\downarrow$ & LPIPS $\downarrow$ \\ \hline\hline
\multirow{5}{*}{\begin{tabular}[c]{@{}c@{}}LS3D-W\\ balanced\\~\cite{LS3D_2017}\end{tabular}}
        & SCGAN-SA    & 25.54 & 2.10$\pm$0.10 & \textbf{0.073}  \\
        & SCGAN-SE  & 23.47 & 2.19$\pm$0.08 & \textbf{0.073} \\
        & SCGAN-SD    & \textbf{23.33} & \textbf{1.39$\pm$0.08} & \textbf{\textcolor{red}{0.068}}  \\
        & SCGAN-SM & \textbf{\textcolor{blue}{23.12}} & \textbf{\textcolor{blue}{1.28$\pm$0.06}} & 0.077  \\
        & SCGAN       & \textbf{\textcolor{red}{22.55}} & \textbf{\textcolor{red}{1.26$\pm$0.06}} & \textbf{\textcolor{red}{0.068}} \\ \hline
\multirow{5}{*}{\begin{tabular}[c]{@{}c@{}}FFHQ\\ \cite{StyleGAN_cvpr2019}\end{tabular}}
        & SCGAN-SA    & \textbf{10.65} & \textbf{\textcolor{blue}{1.05$\pm$0.05}} & \textbf{0.200}  \\
        & SCGAN-SE  & \textbf{\textcolor{blue}{10.59}} & 1.64$\pm$0.07 & 0.209  \\
        & SCGAN-SD    & 10.96 & \textbf{1.11$\pm$0.06} & \textbf{\textcolor{blue}{0.198}}  \\
        & SCGAN-SM & 11.57 & 1.31$\pm$0.07 & 0.219  \\
        & SCGAN       & \textbf{\textcolor{red}{9.06}} & \textbf{\textcolor{red}{0.94$\pm$0.05}} & \textbf{\textcolor{red}{0.197}}  \\ \hline\hline
Dataset & Variant & FID $\downarrow$ & KID $\downarrow$ & NIQE $\downarrow$ \\ \hline\hline
\multirow{5}{*}{\begin{tabular}[c]{@{}c@{}}Widerface\\ \cite{widerface_cvpr2016}\end{tabular}}
        & SCGAN-SA    & 15.36 & 1.45$\pm$0.08 & \textbf{6.6553}  \\
        & SCGAN-SE  & \textbf{14.30} & 1.64$\pm$0.09 & \textbf{\textcolor{blue}{6.6534}}  \\
        & SCGAN-SD    & \textbf{\textcolor{blue}{13.61}} & \textbf{\textcolor{blue}{1.11$\pm$0.05}} & 6.6741 \\
        & SCGAN-SM & 14.58 & \textbf{1.23$\pm$0.07} & 6.7038 \\
        & SCGAN       & \textbf{\textcolor{red}{13.32}} & \textbf{\textcolor{red}{1.08$\pm$0.05}} & \textbf{\textcolor{red}{6.6192}}  \\ \hline
\multirow{5}{*}{\begin{tabular}[c]{@{}c@{}}WebFace \end{tabular}}
        & SCGAN-SA    & 21.77 & 1.86$\pm$0.08 & \textbf{6.6563}  \\
        & SCGAN-SE  & 22.04 & 2.14$\pm$0.08 & \textbf{\textcolor{blue}{6.5968}}  \\
        & SCGAN-SD    & \textbf{\textcolor{blue}{21.49}} & \textbf{\textcolor{blue}{1.43$\pm$0.07}} & 6.6875 \\
        & SCGAN-SM & \textbf{21.69} & \textbf{1.70$\pm$0.07} & 6.7028 \\
        & SCGAN       & \textbf{\textcolor{red}{21.06}} & \textbf{\textcolor{red}{1.39$\pm$0.06}} & \textbf{\textcolor{red}{6.5835}} \\ \hline
\end{tabular}
\label{tab:ablationshare}
\vspace{-7mm}
\end{table}

\begin{table}[t]
\centering
  \footnotesize
  \renewcommand{\arraystretch}{1.2}
  \setlength\tabcolsep{3pt}
  \caption{\small \textbf{Quantitative results on two synthetic and two real-world datasets by our SCGAN and its variants with different loss functions}.
  The best, second best, and third best results are highlighted in \textbf{\textcolor{red}{red}}, \textbf{\textcolor{blue}{blue}} and \textbf{bold}, respectively.}
  \vspace{-3mm}
\begin{tabular}{c||c||c|c|c}
\Xhline{1.6pt}
Dataset & Variant & FID $\downarrow$ & KID $\downarrow$ & LPIPS $\downarrow$ \\ \hline\hline
\multirow{4}{*}{\begin{tabular}[c]{@{}c@{}}LS3D-W\\ balanced\\~\cite{LS3D_2017}\end{tabular}}
        & SCGAN-w/o-AL    
        & 235.60 & \textbf{10.47$\pm$0.10} & \textbf{0.343}  \\
        & SCGAN-w/o-PL  
        & \textbf{185.45} & 24.48$\pm$0.24 & 0.360 \\
        & SCGAN-w/o-CL   
        & \textbf{\textcolor{blue}{33.62}} & \textbf{\textcolor{blue}{4.11$\pm$0.09}} & \textbf{\textcolor{blue}{0.117}}  \\
        & SCGAN 
        & \textbf{\textcolor{red}{22.55}} & \textbf{\textcolor{red}{1.26$\pm$0.06}} & \textbf{\textcolor{red}{0.068}}  \\ \hline\hline
\multirow{4}{*}{\begin{tabular}[c]{@{}c@{}}FFHQ\\ \cite{StyleGAN_cvpr2019}\end{tabular}}
        & SCGAN-w/o-AL    
        & 235.76 & \textbf{11.81$\pm$0.18} & 0.916  \\
        & SCGAN-w/o-PL  
        & \textbf{181.00} & 24.89$\pm$0.35 & \textbf{0.877} \\
        & SCGAN-w/o-CL   
        & \textbf{\textcolor{blue}{24.29}} & \textbf{\textcolor{blue}{4.21$\pm$0.12}} & \textbf{\textcolor{blue}{0.289}}  \\
        & SCGAN 
        & \textbf{\textcolor{red}{9.06}} & \textbf{\textcolor{red}{0.94$\pm$0.05}} & \textbf{\textcolor{red}{0.197}}  \\ \hline\hline
Dataset & Variant & FID $\downarrow$ & KID $\downarrow$ & NIQE $\downarrow$ \\ \hline\hline
\multirow{4}{*}{\begin{tabular}[c]{@{}c@{}}Widerface\\ \cite{widerface_cvpr2016}\end{tabular}}
        & SCGAN-w/o-AL    
        & 233.94 & \textbf{11.00$\pm$0.16} & 7.7086\\
        & SCGAN-w/o-PL  
        & \textbf{186.03} & 24.88$\pm$0.30 & \textbf{6.8927}\\
        & SCGAN-w/o-CL   
        & \textbf{\textcolor{blue}{26.49}} & \textbf{\textcolor{blue}{4.21$\pm$0.12}} & \textbf{\textcolor{blue}{6.7678}}\\
        & SCGAN 
        & \textbf{\textcolor{red}{13.32}} & \textbf{\textcolor{red}{1.08$\pm$0.05}} & \textbf{\textcolor{red}{6.6192}}\\ \hline\hline
\multirow{4}{*}{\begin{tabular}[c]{@{}c@{}}WebFace \end{tabular}}
        & SCGAN-w/o-AL    
        & 239.39 & \textbf{12.49$\pm$0.12} & 7.7209  \\
        & SCGAN-w/o-PL  
        & \textbf{189.44} & 24.55$\pm$0.26 & \textbf{6.9038} \\
        & SCGAN-w/o-CL   
        & \textbf{\textcolor{blue}{31.68}} & \textbf{\textcolor{blue}{4.48$\pm$0.10}} & \textbf{\textcolor{blue}{6.7251}}  \\
        & SCGAN 
        & \textbf{\textcolor{red}{21.06}} & \textbf{\textcolor{red}{1.39$\pm$0.06}} & \textbf{\textcolor{red}{6.5835}}  \\ \hline
\end{tabular}
\label{tab:lossfunctions}
\vspace{-7mm}
\end{table}

% \par
% In order to better understand the roles and performance of the different components of the proposed SCGAN, in this section, we conduct ablation studies by comparing the real-world face SR performance of some variants of SCGAN on Widerface~\cite{widerface_cvpr2016} dataset. The examples of the test results are shown in Figure~\ref{fig:variants}.
% Table~\ref{tab:variants} list the FID, KID and NIQE scores.
\par

% \CheckRmv
% \begin{table}[htbp]% htbp
%   \centering
%   \footnotesize
%   \renewcommand{\arraystretch}{1.2}
%   \setlength\tabcolsep{2.0pt}
%   \caption{\small Comparison (FID, KID and NIQE) of different variants of SCGAN.}
%   \vspace{-3mm}
%   \begin{tabular}{c||c|c|c}
%   \Xhline{1.6pt}
%   Variant & FID $\downarrow$ & KID~$\downarrow$ & NIQE $\downarrow$ \\
%   \hline
%   \hline
%   SCGAN-w/o-D2 & 33.96 & 3.14$\pm$0.11 & \textbf{6.5738}\\
%   SCGAN-w/o-D1 & 20.53 & 3.36$\pm$0.12 & 6.7653\\
%   SCGAN-fc & 16.02 & 1.65$\pm$0.06 & 6.7538\\
%   SCGAN & \textbf{13.32} & \textbf{1.08$\pm$0.05} & \textbf{6.6192}\\
%   \hline
  
%   \hline
%   SCGAN-w/o-AL & 233.54 & 10.82$\pm$0.16 & 7.7086\\
%   SCGAN-w/o-PL & 186.03 & 24.88$\pm$0.30 & 6.8927\\
%   SCGAN-w/o-CL & 26.49 & 4.21$\pm$0.12 & 6.7678\\
%   SCGAN & \textbf{13.32} & \textbf{1.08$\pm$0.05} & \textbf{6.6192}\\
%   \hline
%   \end{tabular}
%   \label{tab:variants}
%   \vspace{-3mm}
% \end{table}

\begin{table*}[]
\centering
  \footnotesize
  \renewcommand{\arraystretch}{1.2}
  \setlength\tabcolsep{8pt}
  \caption{\textbf{Quantitative results on two real-world LR face image datasets by our SCGAN and its 15 more variants with different combinations of adversarial losses}.
  The best, second best, and third best results are highlighted in \textbf{\textcolor{red}{red}}, \textbf{\textcolor{blue}{blue}} and \textbf{bold}, respectively.}
  \vspace{-3mm}
\begin{tabular}{c||c||c|c|c|c|ccc|ccc}
\Xhline{1.6pt}
\multirow{2}{*}{\begin{tabular}[c]{@{}c@{}}Removal\\ number\end{tabular}} &
  \multirow{2}{*}{Variant} &
  \multirow{2}{*}{$l_{adv}^{\mathcal{D}_{L1}}$} &
  \multirow{2}{*}{$l_{adv}^{\mathcal{D}_{L2}}$} &
  \multirow{2}{*}{$l_{adv}^{\mathcal{D}_{H1}}$} &
  \multirow{2}{*}{$l_{adv}^{\mathcal{D}_{H2}}$} &
  \multicolumn{3}{c|}{Widerface~\cite{widerface_cvpr2016}} &
  \multicolumn{3}{c}{WebFace} \\ \cline{7-12} 
 & & & & & &
  \multicolumn{1}{c|}{FID $\downarrow$} &
  \multicolumn{1}{c|}{KID $\downarrow$} &
  \multicolumn{1}{c|}{NIQE $\downarrow$} &

  \multicolumn{1}{c|}{FID $\downarrow$} &
  \multicolumn{1}{c|}{KID $\downarrow$} &
  \multicolumn{1}{c}{NIQE $\downarrow$}  \\ \hline\hline
4 &
  $l_{adv}$-4-1 &
  \ding{55} &
  \ding{55} &
  \ding{55} &
  \ding{55} &
  \multicolumn{1}{c|}{233.94} &
  \multicolumn{1}{c|}{11.00$\pm$0.16} &
  \multicolumn{1}{c|}{7.7086} &
  \multicolumn{1}{c|}{239.39} &
  \multicolumn{1}{c|}{12.49$\pm$0.12} &
  \multicolumn{1}{c}{7.7209}  \\ \hline\hline
\multirow{4}{*}{3} &
  $l_{adv}$-3-1 &
  \ding{55} &
  \ding{55} &
  \ding{55} &
  \checkmark &
  \multicolumn{1}{c|}{31.68} &
  \multicolumn{1}{c|}{3.31$\pm$0.11} &
  \multicolumn{1}{c|}{\textbf{6.7656}} &
  \multicolumn{1}{c|}{41.60} &
  \multicolumn{1}{c|}{4.07$\pm$0.11} &
  \multicolumn{1}{c}{6.7961}  \\ 
 &
  $l_{adv}$-3-2 &
  \ding{55} &
  \ding{55} &
  \checkmark &
  \ding{55} &
  \multicolumn{1}{c|}{222.02} &
  \multicolumn{1}{c|}{9.55$\pm$0.15} &
  \multicolumn{1}{c|}{7.0507} &
  \multicolumn{1}{c|}{228.72} &
  \multicolumn{1}{c|}{10.94$\pm$0.12} &
  \multicolumn{1}{c}{7.0907} \\ 
 &
  $l_{adv}$-3-3 &
  \ding{55} &
  \checkmark &
  \ding{55} &
  \ding{55} &
  \multicolumn{1}{c|}{230.46} &
  \multicolumn{1}{c|}{10.40$\pm$0.15} &
  \multicolumn{1}{c|}{7.1812} &
  \multicolumn{1}{c|}{236.07} &
  \multicolumn{1}{c|}{11.86$\pm$0.12} &
  \multicolumn{1}{c}{7.2225} \\ 
 &
  $l_{adv}$-3-4 &
  \checkmark &
  \ding{55} &
  \ding{55} &
  \ding{55} &
  \multicolumn{1}{c|}{98.33} &
  \multicolumn{1}{c|}{4.98$\pm$0.12} &
  \multicolumn{1}{c|}{6.8866} &
  \multicolumn{1}{c|}{108.11} &
  \multicolumn{1}{c|}{5.65$\pm$0.11} &
  \multicolumn{1}{c}{6.8991} \\ \hline\hline
\multirow{6}{*}{2} &
  $l_{adv}$-2-1 &
  \ding{55} &
  \ding{55} &
  \checkmark &
  \checkmark &
  \multicolumn{1}{c|}{\textbf{\textcolor{blue}{14.00}}} &
  \multicolumn{1}{c|}{\textbf{\textcolor{blue}{1.09$\pm$0.05}}} &
  \multicolumn{1}{c|}{\textbf{\textcolor{blue}{6.6933}}} &
  \multicolumn{1}{c|}{\textbf{\textcolor{blue}{21.51}}} &
  \multicolumn{1}{c|}{\textbf{\textcolor{blue}{1.50$\pm$0.06}}} &
  \multicolumn{1}{c}{\textbf{\textcolor{blue}{6.6727}}} \\ 
 &
  $l_{adv}$-2-2 &
  \ding{55} &
  \checkmark &
  \ding{55} &
  \checkmark &
  \multicolumn{1}{c|}{201.45} &
  \multicolumn{1}{c|}{8.60$\pm$0.14} &
  \multicolumn{1}{c|}{6.9884} &
  \multicolumn{1}{c|}{207.31} &
  \multicolumn{1}{c|}{9.81$\pm$0.11} &
  \multicolumn{1}{c}{7.0166} \\ 
 &
  $l_{adv}$-2-3 &
  \ding{55} &
  \checkmark &
  \checkmark &
  \ding{55} &
  \multicolumn{1}{c|}{213.81} &
  \multicolumn{1}{c|}{9.15$\pm$0.14} &
  \multicolumn{1}{c|}{7.0584} &
  \multicolumn{1}{c|}{220.66} &
  \multicolumn{1}{c|}{10.23$\pm$0.11} &
  \multicolumn{1}{c}{7.0915}  \\ 
 &
  $l_{adv}$-2-4 &
  \checkmark &
  \ding{55} &
  \ding{55} &
  \checkmark &
  \multicolumn{1}{c|}{34.27} &
  \multicolumn{1}{c|}{4.12$\pm$0.12} &
  \multicolumn{1}{c|}{6.8002} &
  \multicolumn{1}{c|}{43.71} &
  \multicolumn{1}{c|}{4.75$\pm$0.12} &
  \multicolumn{1}{c}{6.8045}  \\ 
 &
  $l_{adv}$-2-5 &
  \checkmark &
  \ding{55} &
  \checkmark &
  \ding{55} &
  \multicolumn{1}{c|}{215.76} &
  \multicolumn{1}{c|}{9.65$\pm$0.14} &
  \multicolumn{1}{c|}{7.0541} &
  \multicolumn{1}{c|}{222.01} &
  \multicolumn{1}{c|}{10.29$\pm$0.11} &
  \multicolumn{1}{c}{7.1055} \\ 
 &
  $l_{adv}$-2-6 &
  \checkmark &
  \checkmark &
  \ding{55} &
  \ding{55} &
  \multicolumn{1}{c|}{233.59} &
  \multicolumn{1}{c|}{10.94$\pm$0.16} &
  \multicolumn{1}{c|}{7.2024} &
  \multicolumn{1}{c|}{238.47} &
  \multicolumn{1}{c|}{12.29$\pm$0.12} &
  \multicolumn{1}{c}{7.2139} \\ \hline\hline
\multirow{4}{*}{1} &
  $l_{adv}$-1-1 &
  \ding{55} &
  \checkmark &
  \checkmark &
  \checkmark &
  \multicolumn{1}{c|}{\textbf{17.44}} &
  \multicolumn{1}{c|}{\textbf{1.52$\pm$0.07}} &
  \multicolumn{1}{c|}{6.7663} &
  \multicolumn{1}{c|}{\textbf{24.46}} &
  \multicolumn{1}{c|}{\textbf{1.85$\pm$0.08}} &
  \multicolumn{1}{c}{\textbf{6.7707}}\\ 
 &
  $l_{adv}$-1-2 &
  \checkmark &
  \ding{55} &
  \checkmark &
  \checkmark &
  \multicolumn{1}{c|}{231.55} &
  \multicolumn{1}{c|}{10.97$\pm$0.16} &
  \multicolumn{1}{c|}{7.0690} &
  \multicolumn{1}{c|}{236.46} &
  \multicolumn{1}{c|}{12.42$\pm$0.13} &
  \multicolumn{1}{c}{7.0960}  \\ 
 &
  $l_{adv}$-1-3 &
  \checkmark &
  \checkmark &
  \ding{55} &
  \checkmark &
  \multicolumn{1}{c|}{36.59} &
  \multicolumn{1}{c|}{4.53$\pm$0.12} &
  \multicolumn{1}{c|}{6.8268} &
  \multicolumn{1}{c|}{46.41} &
  \multicolumn{1}{c|}{5.09$\pm$0.12} &
  \multicolumn{1}{c}{6.8242}  \\ 
 &
  $l_{adv}$-1-4 &
  \checkmark &
  \checkmark &
  \checkmark &
  \ding{55} &
  \multicolumn{1}{c|}{208.74} &
  \multicolumn{1}{c|}{8.77$\pm$0.13} &
  \multicolumn{1}{c|}{7.0242} &
  \multicolumn{1}{c|}{217.33} &
  \multicolumn{1}{c|}{9.66$\pm$0.11} &
  \multicolumn{1}{c}{7.0768}  \\ \hline\hline
0 &
  SCGAN &
  \checkmark &
  \checkmark &
  \checkmark &
  \checkmark &
  \multicolumn{1}{c|}{\textbf{\textcolor{red}{13.32}}} &
  \multicolumn{1}{c|}{\textbf{\textcolor{red}{1.08$\pm$0.05}}} &
  \multicolumn{1}{c|}{\textbf{\textcolor{red}{6.6192}}} &
  \multicolumn{1}{c|}{\textbf{\textcolor{red}{21.06}}} &
  \multicolumn{1}{c|}{\textbf{\textcolor{red}{1.39$\pm$0.06}}} &
  \multicolumn{1}{c}{\textbf{\textcolor{red}{6.5835}}}  \\ \hline
\end{tabular}
\label{tab:RMganloss}
\vspace{-6mm}
\end{table*}

\begin{table}[t]
\centering
  \footnotesize
  \renewcommand{\arraystretch}{1.2}
  \setlength\tabcolsep{3pt}
  \caption{\small  \textbf{Quantitative results on two real-world datasets by our SCGAN and its variants with different structures in two degradation branches}.
  The best, second best, and third best results are highlighted in \textbf{\textcolor{red}{red}}, \textbf{\textcolor{blue}{blue}} and \textbf{bold}, respectively.}
  \vspace{-3mm}
\begin{tabular}{c||c||c|c|c}
\Xhline{1.6pt}
Dataset & Variant & FID $\downarrow$ & KID $\downarrow$ & NIQE $\downarrow$ \\ \hline\hline
\multirow{3}{*}{\begin{tabular}[c]{@{}c@{}}Widerface\\ \cite{widerface_cvpr2016}\end{tabular}}
        & SCGAN-$D_{HL}$-6 & \textbf{21.47} & \textbf{1.89$\pm$0.07} & \textbf{6.8726} \\
        & SCGAN-$D_{SL}$-6 & \textbf{\textcolor{blue}{15.00}} & \textbf{\textcolor{blue}{1.34$\pm$0.07}} & \textbf{\textcolor{blue}{6.7011}} \\
        & SCGAN       & \textbf{\textcolor{red}{13.32}} & \textbf{\textcolor{red}{1.08$\pm$0.05}} & \textbf{\textcolor{red}{6.6192}}  \\ \hline
\multirow{3}{*}{\begin{tabular}[c]{@{}c@{}}WebFace \end{tabular}}
        & SCGAN-$D_{HL}$-6 & \textbf{27.53} & \textbf{2.11$\pm$0.06} & \textbf{6.8160} \\
        & SCGAN-$D_{SL}$-6 & \textbf{\textcolor{blue}{21.99}} & \textbf{\textcolor{blue}{1.46$\pm$0.07}} & \textbf{\textcolor{blue}{6.6485}} \\
        & SCGAN       & \textbf{\textcolor{red}{21.06}} & \textbf{\textcolor{red}{1.39$\pm$0.06}} & \textbf{\textcolor{red}{6.5835}} \\ \hline
\end{tabular}
\label{tab:ablationshallow}
\vspace{-3mm}
\end{table}

\begin{table}[t]
\centering
  \footnotesize
  \renewcommand{\arraystretch}{1.2}
  \setlength\tabcolsep{3pt}
  \caption{\small  \textbf{Quantitative results on two real-world datasets by our SCGAN and its variant with two independent restoration branches and a shared degradation branch}.
  The best, second best, and third best results are highlighted in \textbf{\textcolor{red}{red}}, \textbf{\textcolor{blue}{blue}} and \textbf{bold}, respectively.}
  \vspace{-3mm}
\begin{tabular}{c||c||c|c|c}
\Xhline{1.6pt}
Dataset & Variant & FID $\downarrow$ & KID $\downarrow$ & NIQE $\downarrow$ \\ \hline\hline
\multirow{3}{*}{\begin{tabular}[c]{@{}c@{}}Widerface\\ \cite{widerface_cvpr2016}\end{tabular}}
        & SCGAN-2SR ($\mathcal{R}_{LS}$) & \textbf{397.81} & \textbf{51.88$\pm$0.27} & \textbf{7.8605} \\
        & SCGAN-2SR ($\mathcal{R}_{RS}$) & \textbf{\textcolor{blue}{91.86}} & \textbf{\textcolor{blue}{9.55$\pm$0.18}} & \textbf{\textcolor{red}{6.4488}} \\
        & SCGAN       & \textbf{\textcolor{red}{13.32}} & \textbf{\textcolor{red}{1.08$\pm$0.05}} & \textbf{\textcolor{blue}{6.6192}}  \\ \hline
\multirow{3}{*}{\begin{tabular}[c]{@{}c@{}}WebFace \end{tabular}}
        & SCGAN-2SR ($\mathcal{R}_{LS}$) & \textbf{375.92} & \textbf{51.63$\pm$0.25} & \textbf{7.7761} \\
        & SCGAN-2SR ($\mathcal{R}_{RS}$) & \textbf{\textcolor{blue}{87.29}} & \textbf{\textcolor{blue}{8.59$\pm$0.14}} & \textbf{\textcolor{red}{6.3390}} \\
        & SCGAN       & \textbf{\textcolor{red}{21.06}} & \textbf{\textcolor{red}{1.39$\pm$0.06}} & \textbf{\textcolor{blue}{6.5835}} \\ \hline
\end{tabular}
\label{tab:ablationshallow}
\vspace{-3mm}
\end{table}

\begin{table}[t]
\centering
  \footnotesize
  \renewcommand{\arraystretch}{1.2}
  \setlength\tabcolsep{3pt}
  \caption{\small  \textbf{Quantitative results on two synthetic and two real-world datasets by our SCGAN and its variants with different weights of different loss functions.}.
  The best, second best, and third best results are highlighted in \textbf{\textcolor{red}{red}}, \textbf{\textcolor{blue}{blue}} and \textbf{bold}, respectively.}
  \vspace{-3mm}
\begin{tabular}{c||c||c||c|c|c}
\Xhline{1.6pt}
Dataset & $\beta$ & $\gamma$ & FID $\downarrow$ & KID $\downarrow$ & NIQE $\downarrow$ \\ \hline\hline
% \multirow{5}{*}{\begin{tabular}[c]{@{}c@{}}LS3D-W\\ balanced\\~\cite{LS3D_2017}\end{tabular}}
%         & SCGAN-SA    & 25.54 & 2.10$\pm$0.10 & \textbf{0.073}  \\
%         & SCGAN-SE  & 23.47 & 2.19$\pm$0.08 & \textbf{0.073} \\
%         & SCGAN-SD    & \textbf{23.33} & \textbf{1.39$\pm$0.08} & \textbf{\textcolor{red}{0.068}}  \\
%         & SCGAN-SM & \textbf{\textcolor{blue}{23.12}} & \textbf{\textcolor{blue}{1.28$\pm$0.06}} & 0.077  \\
%         & SCGAN       & \textbf{\textcolor{red}{22.55}} & \textbf{\textcolor{red}{1.26$\pm$0.06}} & \textbf{\textcolor{red}{0.068}} \\ \hline
% \multirow{5}{*}{\begin{tabular}[c]{@{}c@{}}FFHQ\\ \cite{StyleGAN_cvpr2019}\end{tabular}}
%         & SCGAN-SA    & \textbf{10.65} & \textbf{\textcolor{blue}{1.05$\pm$0.05}} & \textbf{0.200}  \\
%         & SCGAN-SE  & \textbf{\textcolor{blue}{10.59}} & 1.64$\pm$0.07 & 0.209  \\
%         & SCGAN-SD    & 10.96 & \textbf{1.11$\pm$0.06} & \textbf{\textcolor{blue}{0.198}}  \\
%         & SCGAN-SM & 11.57 & 1.31$\pm$0.07 & 0.219  \\
%         & SCGAN       & \textbf{\textcolor{red}{9.06}} & \textbf{\textcolor{red}{0.94$\pm$0.05}} & \textbf{\textcolor{red}{0.197}}  \\ \hline\hline
% Dataset & Variant & FID $\downarrow$ & KID $\downarrow$ & NIQE $\downarrow$ \\ \hline\hline
\multirow{9}{*}{\begin{tabular}[c]{@{}c@{}}Widerface\\ \cite{widerface_cvpr2016}\end{tabular}}
        & 1 & 0.05 & 387.24 & 48.47$\pm$0.23 & 7.3197 \\
        & 0.5 & 0.05 & 56.55 & 5.01$\pm$0.13 & \textbf{6.6583} \\
        & 0.1 & 0.05 & \textbf{15.90} & \textbf{1.34$\pm$0.07} & \textbf{\textcolor{red}{6.6016}} \\
        & 0.01 & 0.05 & 18.78 & 1.86$\pm$0.08 & 6.6731 \\
        & 0.05 & 1 & 27.56 & 2.76$\pm$0.11 & 6.7708 \\
        & 0.05 & 0.5 & 23.82 & 2.20$\pm$0.09 & 6.7727 \\
        & 0.05 & 0.1 & \textbf{\textcolor{blue}{15.51}} & \textbf{\textcolor{blue}{1.28$\pm$0.06}} & 6.7463 \\
        & 0.05 & 0.01 & 17.21 & 1.67$\pm$0.08 & 6.7041 \\
        & 0.05 & 0.05       & \textbf{\textcolor{red}{13.32}} & \textbf{\textcolor{red}{1.08$\pm$0.05}} & \textbf{\textcolor{blue}{6.6192}}  \\ \hline
\multirow{9}{*}{\begin{tabular}[c]{@{}c@{}}WebFace \end{tabular}}
        & 1 & 0.05 & 366.09 & 48.24$\pm$0.21 & 7.4073 \\
        & 0.5 & 0.05 & 30.62 & 5.08$\pm$0.11 & 6.6584 \\
        & 0.1 & 0.05 & \textbf{\textcolor{blue}{23.41}} & \textbf{1.73$\pm$0.07} & \textbf{\textcolor{red}{6.5631}} \\
        & 0.01 & 0.05 & 30.00 & 2.60$\pm$0.08 & 6.6624 \\
        & 0.05 & 1 & 33.39 & 3.35$\pm$0.10 & 6.7788 \\
        & 0.05 & 0.5 & 31.99 & 2.71$\pm$0.08 & 6.7178 \\
        & 0.05 & 0.1 & \textbf{25.31} & \textbf{\textcolor{blue}{1.68$\pm$0.06}} & 6.7091 \\
        & 0.05 & 0.01 & 25.94 & 2.07$\pm$0.08 & \textbf{6.6479} \\
        & 0.05 & 0.05       & \textbf{\textcolor{red}{21.06}} & \textbf{\textcolor{red}{1.39$\pm$0.06}} & \textbf{\textcolor{blue}{6.5835}} \\ \hline
        
\end{tabular}
\label{tab:ablationweights}
\vspace{-3mm}
\end{table}

% 1和2整合到一起
\noindent
\textbf{a) How the semi-cycled architecture benefits our SCGAN on real-world face SR?}
To answer this question, we develop three variants of our SCGAN.
1) We remove the real-world HR face degradation branch $\mathcal{D}_{SL}$ introduced in \S\ref{sec:branch3}, and only train the forward cycle-consistency reconstruction process ``$\mathcal{D}_{HL}\rightarrow\mathcal{R}_{LS}$''.
This variant is denoted as ``SCGAN-w/o-$\mathcal{D}_{SL}$''.
2) We remove the synthetic degradation branch $\mathcal{D}_{HL}$ introduced in \S\ref{sec:branch1}, and only train the backward cycle-consistency reconstruction process ``$\mathcal{R}_{LS}\rightarrow\mathcal{D}_{SL}$''.
This variant is denoted as ``SCGAN-w/o-$\mathcal{D}_{HL}$''.
3) We share the parameters of synthetic HR face degradation branch $\mathcal{D}_{HL}$ and real-world one $\mathcal{D}_{SL}$, and jointly train the forward cycle-consistency reconstruction process $\mathcal{D}_{HL}\rightarrow\mathcal{R}_{LS}$ as well as the backward one $\mathcal{R}_{LS}\rightarrow\mathcal{D}_{SL}$.
This variant is denoted as ``SCGAN-fc''.
% We train these variants under the same settings as our SCGAN and evaluate them on the four test sets.
%
The quantitative results are listed in Table~\ref{tab:variants}.
We observe that our SCGAN achieves better results in term of FID, KID, and LPIPS, with comparable NIQE results, than the other variants.
This demonstrates that our semi-cycled architecture really benefits the real-world face SR task.

\noindent
\textbf{b) Whether to share parameters or not in the two degradation branches in our SCGAN?}
Here, we study whether to share parameters or not in the two degradation branches $\mathcal{D}_{HL}$ and $\mathcal{D}_{SL}$ of our semi-cycled SCGAN.
To this end, we design four other variants of our SCGAN with degradation branches $\mathcal{D}_{HL}$ sharing partial parameters (except for the input head and output head).
1) We share the parameters of all layers except the input and output heads (denoted as ``In'' and ``Out'' in Figure~\ref{fig:netconfig}) in $\mathcal{D}_{HL}$ and $\mathcal{D}_{SL}$, and denote this variant as ``SCGAN-SA''.
2) We share the parameters of the encoder layers in $\mathcal{D}_{HL}$ and $\mathcal{D}_{SL}$, and denote this variant as ``SCGAN-SE''.
3) We share the parameters of the decoder layers in $\mathcal{D}_{HL}$ and $\mathcal{D}_{SL}$, and denote this variant as ``SCGAN-SD''.
4) We share the parameters of the middle part, \ie, the last two groups of Resblocks in the encoder and the first two groups in the decoder, in $\mathcal{D}_{HL}$ and $\mathcal{D}_{SL}$.
We denote this variant as ``SCGAN-SM''.
The objective results are listed in Table~\ref{tab:ablationshare}.
One can see that our SCGAN with two independent degradation branch achieves the best results among all these variants in terms of all four objective metrics. 
% Second, SCGAN-SD, the variant of shared decoder layers parameters, obtains results closest to our SCGAN, so we can conclude that the domain gap has a more pronounced detrimental effect on the encoding stage of the degradation process.

\noindent
\textbf{c) How different loss functions (\ie, adversarial loss, pixel loss and cycle consistency loss) contribute to our
SCGAN on face SR?}
To understand the role of different loss functions, we design three variants of our SCGAN:
1) we remove all adversarial losses in our SCGAN, and denote this variant as ``SCGAN-w/o-AL'';
2) we remove all pixel losses in our SCGAN, and denote this variant as ``SCGAN-w/o-PL'';
3) we remove all cycle-consistency losses in our SCGAN, and denote this variant as ``SCGAN-w/o-CL''.
The results of FID, KID and NIQE listed in Table~\ref{tab:lossfunctions} show that our SCGAN without either loss function achieves inferior performance to the original SCGAN.
The visual comparison results on Widerface~\cite{widerface_cvpr2016} are shown in Figure~\ref{fig:variants}.
We observe that the variant ``SCGAN-w/o-AL'' fails to recover well the face details, and the variant ``SCGAN-w/o-PL'' could not guarantee the contextual consistency with the input real-world LR face images, while the variant ``SCGAN-w/o-CL'' hard to preserve structural consistency on identity.
On the contrary, by integrating all three loss functions, our SCGAN recovers well the contextual and detailed information to preserve the face identity.
These demonstrate that the adversarial loss is mainly used to recover details, and the pixel loss is mainly used to preserve the contextual information, while the cycle-consistency loss is mainly used to keep the structural consistency.

\begin{figure*}[t] 
\centering
	\begin{overpic}[width=\textwidth]{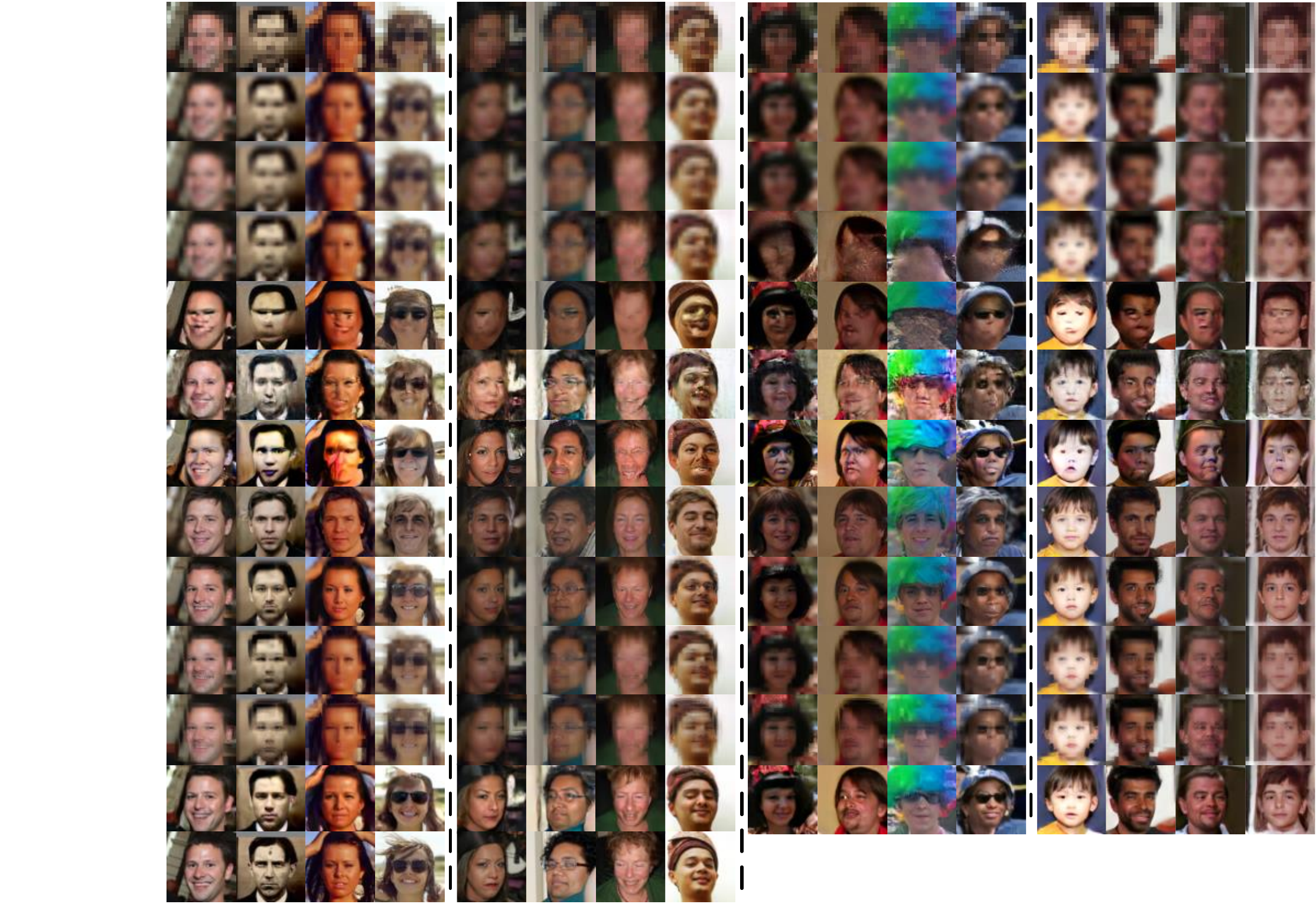}
	\put(2.2,65.5){\textbf{\scriptsize{LR face images}}}
	\put(5.6,60.8){\textbf{\scriptsize{Bicubic}}}
	\put(3.6,59.4){\textbf{\scriptsize{Interpolation}}}
	\put(3.8,54.8){\textbf{\scriptsize{DFDNet~\cite{DFDNet_eccv2018}}}}
	\put(1.8,49.4){\textbf{\scriptsize{HifaceGAN~\cite{hifacegan_acmmm2020}}}}
	\put(-0.2,44.4){\textbf{\scriptsize{Real-ESRGAN~\cite{realesrgan_cvpr2021}}}}
	\put(2.2,39.0){\textbf{\scriptsize{CycleGAN~\cite{cyclegan2017}}}}
	\put(3.6,34){\textbf{\scriptsize{LRGAN~\cite{LRGAN}}}}
	\put(4.1,28.4){\textbf{\scriptsize{PULSE~\cite{pulse_cvpr2020}}}}
	\put(2.8,23.2){\textbf{\scriptsize{GFPGAN~\cite{GFPGAN2021}}}}
	\put(3.9,18.0){\textbf{\scriptsize{GCFSR~\cite{GCFSR_CVPR2022}}}}
	\put(-0.4,12.8){\textbf{\scriptsize{RestoreFormer~\cite{RestoreFormer_CVPR2022}}}}
	\put(2.6,7.4){\textbf{\scriptsize{SCGAN (Ours)}}}
    \put(2.8,2.2){\textbf{\scriptsize{Ground Truth}}}
    \put(69.0,2.2){\textbf{\large{No Ground-Truth}}}
    \put(16,69.4){\textbf{\footnotesize{LS3D-W balanced~\cite{LS3D_2017}}}}
    \put(42,69.4){\textbf{\footnotesize{FFHQ~\cite{StyleGAN_cvpr2019}}}}
    \put(62.6,69.4){\textbf{\footnotesize{Widerface~\cite{widerface_cvpr2016}}}}
    \put(86.4,69.4){\textbf{\footnotesize{WebFace}}}
%   \put(55,10){\includegraphics[width=\textwidth]{Imgs/tsne.pdf}}
    \end{overpic}

\caption{\textbf{Comparison of visual quality by our SCGAN and other face SR methods} on LS3D-W balanced~\cite{LS3D_2017}, FFHQ~\cite{StyleGAN_cvpr2019}, Widerface~\cite{widerface_cvpr2016}, and WebFace datasets, respectively (from left to right).
Please zoom in for better view.}
\label{fig:fsrandrfsr}
\vspace{-3mm}
\end{figure*}

\noindent
\textbf{d) How different combinations of adversarial losses influence our SCGAN?}
Our SCGAN has 4 adversarial losses.
To study this problem, we design 15 more variants of our SCGAN in 4 categories, according to the number of removed adversarial losses.
The variants are denoted as ``$l_{adv}$-a-b'', where ``a'' represents the number of removed adversarial losses, and ``b'' represents the possible combination of the remaining 4-a adversarial losses.
The details of the variants and the objective results are summarized in Table~\ref{tab:RMganloss}.
We have four main observations:
1) With all the 4 adversarial losses, our SCGAN achieves better results than the other 15 variants.
This demonstrates the essential role of every adversarial loss in our SCGAN for promising face SR performance.
2) By removing one adversarial loss, our SCGAN without $\mathcal{D}_{L2}$ or $\mathcal{D}_{H2}$ degrades greatly in terms of all four evaluation metrics.
This reveals the dominate role of the adversarial losses in the backward cycle-consistency learning process for effective real-world LR face restoration.
% Besides, our SCGAN without $\mathcal{D}_{L1}$ (``$l_{adv}$-1-1'') still achieves satisfactory face SR results.
%
3) By removing two adversarial losses, our SCGAN without $\mathcal{D}_{L1}$ and $\mathcal{D}_{L2}$ (``$l_{adv}$-2-1'') achieves slightly inferior results than our original SCGAN.
This shows the essential role of high-quality HR face images on the guidance of learning an effective LR face restoration branch.
% \textcolor{red}{Besides, our SCGAN without $\mathcal{D}_{L1}$ and $\mathcal{D}_{L2}$ (``$l_{adv}$-2-1'') obtains even better objective results than our SCGAN without $\mathcal{D}_{L1}$ (``$l_{adv}$-1-1''). The reason is that }.
%
4) By removing three adversarial losses, the variant ``$l_{adv}$-3-1'' performs better than the other three variants of ``$l_{adv}$-3-2'', ``$l_{adv}$-3-3'', or ``$l_{adv}$-3-4''.
This shows that the adversarial loss in $\mathcal{D}_{H2}$ plays a dominant role in optimizing the real-world LR face restoration branch.

\noindent
\textbf{e) How different structures of the degradation branch influence our SCGAN?}
To this end, we conducted experiments to explore the impact of degradation branch architectures on the performance of our SCGAN. Specifically, we designed two variants of our SCGAN with different degradation structures: 1) ``SCGAN-$D_{HL}$-6'': the number of residual blocks in the encoder-decoder architecture of the synthetic degradation branch $D_{HL}$ of our SCGAN is reduced from 12 to 6; 2) ``SCGAN-$D_{SL}$-6'': the number of residual blocks in the encoder-decoder architecture of the real-world HR face degradation branch $D_{SL}$ of our SCGAN is reduced from 12 to 6.
In Table~\ref{tab:ablationshallow}, we list the quantitative results on two real-world datasets Widerface~\cite{widerface_cvpr2016} and WebFace by the two variants ``SCGAN-$D_{HL}$-6'' and ``SCGAN-$D_{SL}$-6'', as well as our SCGAN.
One can see that the variants of ``SCGAN-$D_{HL}$-6'' and ``SCGAN-$D_{SL}$-6'' achieve lower FID, KID, and NIQE scores than the proposed SCGAN.
This indicates that our SCGAN performs better when the two degradation branches both have 12 residual blocks than those with only 6 blocks in the synthetic (or real-world) HR face degradation branch.

\noindent
\textbf{f) How about using two independent restoration branches with a shared degradation branch?}
To answer this question, we performed experiments by designing our SCGAN with a shared degradation branch and two independent restoration branches.
We denote this varaint as ``SCGAN-2SR'', it contains two independent restoration branches, denoted as $\mathcal{R}_{LS}$ and $\mathcal{R}_{RS}$.
Please note that, since this variant ``SCGAN-2SR'' contains two restoration branches $\mathcal{R}_{LS}$ and $\mathcal{R}_{RS}$, here we performed independent face image super-resolution test by $\mathcal{R}_{LS}$ and $\mathcal{R}_{RS}$, respectively.
In Table~\ref{tab:ablationshallow}, we list the quantitative results on two real-world datasets Widerface~\cite{widerface_cvpr2016} and WebFace by the restoration branch $\mathcal{R}_{LS}$ or $\mathcal{R}_{RS}$ in the variant ``SCGAN-2SR'', as well as our SCGAN.
It can be seen that, our SCGAN achieves better results than the restoration branches $\mathcal{R}_{LS}$ and $\mathcal{R}_{RS}$ of the variant ``SCGAN-2SR''.
This demonstrates that using two independent SR branches with one shared degradation branch largely degrades the performance of our SCGAN on real-world face image super-resolution task.

\noindent
\textbf{g) How to determine the weights of different loss functions?}
The proposed SCGAN has four weights ($\alpha$, $\beta$, $\theta$, and $\gamma$) for different loss functions.
To determine these parameters, we have conducted more ablation studies with different weights $\beta=0.01, 0.1, 0.5, 1$ and $\gamma=0.01, 0.1, 0.5, 1$ by fixing one parameter as $0.05$.
The quantitative results presented in Table~\ref{tab:ablationweights} show that, although our SCGAN achieves the best NIQE score when $\beta=0.1$ and $\gamma=0.05$, our SCGAN obtains the best FID and KID scores on both datasets when $\beta=0.05$ and $\gamma=0.05$.
Overall, we set $\beta=0.05$ and $\gamma=0.05$.

% \par
% \textbf{6) How does the pixel loss influence the performance of our SCGAN model?}
% We remove the pixel loss and train with the same training strategy, denote by SCGAN-w/o-PL. The test results are shown in Figure~\ref{fig:variants}. One can see that, the obtained results tend to be the same face, and we believe that the generated face image cannot ensure consistency in content with the original LR face image without the constraint of pixel loss. Moreover, it can be seen from Table~\ref{tab:variants} that the obtained results cannot guarantee good visual quality. Furthermore, we retrain our SCGAN with different pixel losses, and the objective evaluation metrics of the test results are shown in Table~\ref{tab:pixloss}.

% \par
% \textbf{7) How does the cycle consistency loss influence the performance of our SCGAN model?} We removed the cycle consistency loss and retrained the variant, denoted by SCGAN-w/o-CL. The test results are shown in Figure~\ref{fig:variants}. It can be seen that without cycle consistency as a constraint, the results obtained by this variant neither have high visual quality nor well guarantee the consistency of face structure.

\vspace{-3mm}

\subsection{Comparisons with State-of-the-art Methods}
\label{sec:comparisonwithsota}
Here, we compare our SCGAN with the state-of-the-art methods on both synthetic and real-world $\times4$ face SR tasks ($16\times16$ LR face images to $64\times64$ HR ones).
To comprehensively evaluate the performance of different methods on face SR, we perform face SR with three different degradation settings: 1) \textsl{simple} degradation with randomly bilinear or bicubic downsampling; 2) \textsl{complex} degradation with blur kernel, downsampling, synthetic noise, and JPEG compression; and 3) \textsl{real-world} unknown degradation.

\noindent
\textbf{Comparison methods}.
We compare our SCGAN with Bicubic Interpolation and other state-of-the-art methods, such as DFDNet~\cite{DFDNet_eccv2018}, HifaceGAN~\cite{hifacegan_acmmm2020}, Real-ESRGAN~\cite{realesrgan_cvpr2021}, GFPGAN~\cite{GFPGAN2021}, LRGAN~\cite{LRGAN}, PULSE~\cite{pulse_cvpr2020}, GCFSR~\cite{GCFSR_CVPR2022}, and RestoreFormer~\cite{RestoreFormer_CVPR2022}.
We also compare our SCGAN with the fully-cycled CycleGAN~\cite{cyclegan2017} to verify the effectiveness of our semi-cycled SCGAN on face super-resolution.
Here, DFDNet~\cite{DFDNet_eccv2018} firstly detects and crops out faces using a face detector and then performs face super-resolution on the cropped images, and would be limited by the accuracy of employed face detector.
The PULSE~\cite{pulse_cvpr2020} utilized a pre-trained StyleGAN~\cite{StyleGAN_cvpr2019}, and thus can only generate HR images of size $1024\times1024$.
For a fair comparison, we resize its results to the same size (\eg, $64\times64$) as those obtained by the comparison methods.

\begin{table*}[t]
\centering
  \footnotesize
  \renewcommand{\arraystretch}{1.2}
  \setlength\tabcolsep{3.8pt}
  \caption{\small  \textbf{Quantitative results of our SCGAN and other state-of-the-art methods on two synthetic and two real-world datasets.} The best, second best, and third best results are highlighted in \textbf{\textcolor{red}{red}}, \textbf{\textcolor{blue}{blue}} and \textbf{bold}, respectively.}
  \vspace{-3mm}
  \resizebox{!}{30mm}{
\begin{tabular}{c||c||c|c|c|c|c|c|c|c|c|c|c}
\Xhline{1.6pt}
\multirow{2}{*}{Dataset} &
  \multirow{2}{*}{Metric} &
  \multirow{2}{*}{\begin{tabular}[c]{@{}c@{}}Bicubic\\ Interpolation\end{tabular}} &
  \multirow{2}{*}{\begin{tabular}[c]{@{}c@{}}DFDNet\\ \cite{DFDNet_eccv2018}\end{tabular}} &
  \multirow{2}{*}{\begin{tabular}[c]{@{}c@{}}HifaceGAN\\ \cite{hifacegan_acmmm2020}\end{tabular}} &
  \multirow{2}{*}{\begin{tabular}[c]{@{}c@{}}Real-ESRGAN\\ \cite{realesrgan_cvpr2021}\end{tabular}} &
  \multirow{2}{*}{\begin{tabular}[c]{@{}c@{}}CycleGAN\\ \cite{cyclegan2017}\end{tabular}} &
  \multirow{2}{*}{\begin{tabular}[c]{@{}c@{}}GFPGAN\\ \cite{GFPGAN2021}\end{tabular}} &
  \multirow{2}{*}{\begin{tabular}[c]{@{}c@{}}LRGAN\\ \cite{LRGAN}\end{tabular}} &
  \multirow{2}{*}{\begin{tabular}[c]{@{}c@{}}PULSE\\ \cite{pulse_cvpr2020}\end{tabular}} &
  \multirow{2}{*}{\begin{tabular}[c]{@{}c@{}}GCFSR\\ \cite{GCFSR_CVPR2022}\end{tabular}} &
  \multirow{2}{*}{\begin{tabular}[c]{@{}c@{}}RestoreFormer\\ \cite{RestoreFormer_CVPR2022}\end{tabular}} &
  \multirow{2}{*}{\begin{tabular}[c]{@{}c@{}}\textcolor{red}{SCGAN}\\ (Ours)\end{tabular}} \\
   &     &     &     &     &     &     &     &     &     &     &     &     \\ \hline\hline
\multirow{3}{*}{\begin{tabular}[c]{@{}c@{}}\emph{LS3D-W}\\ \emph{balanced}\\ \cite{LS3D_2017} \end{tabular}} 
& FID$\downarrow$      & 274.78 & 277.38 & 238.75 & 57.20 & 51.04 & 51.02 & \textbf{33.67} & \textbf{\textcolor{blue}{33.65}} &70.58 &72.89 & \textbf{\textcolor{red}{22.55}} \\ 
& KID$\downarrow$      & 15.31$\pm$0.11 & 16.60$\pm$0.11 & 13.56$\pm$0.12 & \textbf{\textcolor{blue}{3.08$\pm$0.07}} & 5.81$\pm$0.11 & 4.45$\pm$0.11 & 3.95$\pm$0.10 & \textbf{3.49$\pm$0.16} &5.37$\pm$0.10 &5.45$\pm$0.11  & \textbf{\textcolor{red}{1.26$\pm$0.06}} \\ 
& LPIPS$\downarrow$    & 0.343 & 0.387 & 0.268 & 0.114 & \textbf{\textcolor{blue}{0.086}} & 0.094 & 0.118 & \textbf{0.106} &0.125 &0.142  & \textbf{\textcolor{red}{0.068}} \\ \hline\hline
\multirow{3}{*}{\begin{tabular}[c]{@{}c@{}}\emph{FFHQ}\\ \cite{StyleGAN_cvpr2019} \end{tabular}}              
& FID$\downarrow$      & 277.19 & 285.93 & 241.69 & 43.75 & 36.22 & 43.86 & \textbf{\textcolor{blue}{18.39}} & \textbf{27.81} &42.57 &46.72  & \textbf{\textcolor{red}{9.06}} \\ 
& KID$\downarrow$      & 16.32$\pm$0.22 & 23.84$\pm$0.28 & 14.52$\pm$0.19 & \textbf{\textcolor{blue}{3.13$\pm$0.10}} & 6.14$\pm$0.15 & 3.88$\pm$0.12 & 3.22$\pm$0.13 & 3.48$\pm$0.19 &4.00$\pm$1.13 &3.84$\pm$1.18  & \textbf{\textcolor{red}{0.94$\pm$0.05}} \\ 
& LPIPS$\downarrow$    & 0.927 & 0.959 & 0.719 & 0.336 & \textbf{\textcolor{blue}{0.234}} & \textbf{0.299} & 0.302 & 0.329 &0.317 &0.336  & \textbf{\textcolor{red}{0.197}} \\ \hline\hline
\multirow{3}{*}{\begin{tabular}[c]{@{}c@{}}\emph{Widerface}\\ \cite{widerface_cvpr2016}\end{tabular}}         
& FID$\downarrow$      & 270.69         & 271.74 & 157.34 & 41.73 & 39.55 & 59.34 & \textbf{\textcolor{blue}{19.35}} & \textbf{28.27} &54.69 &59.08  & \textbf{\textcolor{red}{13.32}} \\ 
& KID$\downarrow$      & 16.23$\pm$0.21 & 17.34$\pm$0.22 & 17.54$\pm$0.24 & \textbf{\textcolor{blue}{2.77$\pm$0.10}} & 6.26$\pm$0.12 & 3.79$\pm$0.12 & \textbf{3.20$\pm$0.14} & 3.36$\pm$0.17 &4.80$\pm$0.12 &4.83$\pm$0.13  & \textbf{\textcolor{red}{1.08$\pm$0.05}} \\  
& NIQE$\downarrow$     & 18.8702        & 7.9981 & \textbf{\textcolor{red}{5.2294}} & 6.7595 & 6.8953 & 6.9437 & 6.6446 & 6.7473 &\textbf{\textcolor{blue}{5.9455}} &6.7386  & \textbf{6.6192} \\  \hline\hline
\multirow{3}{*}{\emph{WebFace}}                                                               
& FID$\downarrow$      & 273.96         & 274.49 & 238.10 & 51.57 & 44.94 & 88.71 & \textbf{\textcolor{blue}{26.49}} & \textbf{36.02} &70.57 &76.51  & \textbf{\textcolor{red}{21.06}} \\ 
& KID$\downarrow$      & 18.02$\pm$0.15 & 19.08$\pm$0.15 & 14.46$\pm$0.15 & \textbf{3.62$\pm$0.09} & 6.54$\pm$0.11 & 6.45$\pm$0.11 & 4.45$\pm$0.11 & \textbf{\textcolor{blue}{3.60$\pm$0.16}} &5.89$\pm$0.11 &5.70$\pm$0.10  & \textbf{\textcolor{red}{1.39$\pm$0.06}} \\ 
& NIQE$\downarrow$     & 7.4657         & 7.7320 & \textbf{\textcolor{blue}{6.0178}} & 6.6773 & 6.8132 & 6.8841 & \textbf{6.4868} & 6.7039 &\textbf{\textcolor{red}{5.9873}}&6.8277  & 6.5835 \\  \hline\hline
\multirow{3}{*}{\begin{tabular}[c]{@{}c@{}}\emph{DroneSURF}\\ \cite{dronesurf_2019}\end{tabular}}                                                           
& FID$\downarrow$      & 173.06         & 247.77 & 153.01 & \textbf{58.47} & 87.81 & 114.05 & 59.25 & \textbf{\textcolor{red}{27.91}} &116.05 & 147.58  & \textbf{\textcolor{blue}{48.08}} \\ 
& KID$\downarrow$      & 21.27$\pm$1.50         & 35.54$\pm$1.58 & 20.33$\pm$1.59 & \textbf{4.96$\pm$1.09} & 8.94$\pm$1.52 & 15.03$\pm$1.25 & 7.13$\pm$1.25 & \textbf{\textcolor{red}{4.04$\pm$1.36}} &14.61$\pm$1.26 &15.85$\pm$1.35  & \textbf{\textcolor{blue}{4.62$\pm$1.14}} \\ 
& NIQE$\downarrow$     & 6.4464         & 7.9855 & 5.8811 & \textbf{\textcolor{red}{4.9336}} & 7.1707 & \textbf{5.7217} & 6.6369 & 6.8865 &6.5101 &6.5475  & \textbf{\textcolor{blue}{5.1488}} \\  \hline\hline 
\end{tabular}}
\label{tab:FaceSRcomparison}
\vspace{-6mm}
\end{table*}

\noindent
\textbf{Face SR on \textsl{simple} degradation}.
Here, the \textsl{simple} degradation is performed by random bilinear or bicubic downsampling on the 1,000 LR face images in LS3D-W balanced dataset~\cite{LS3D_2017} as the test set, as described in \S\ref{sec:dataset}, and we evaluate the performance of different methods on it.
%
% We evaluate the performance of different methods on a synthetic LS3D-W balanced dataset~\cite{LS3D_2017}, as described in \S\ref{sec:dataset}.
%
For a fair comparison, all the comparison methods are re-trained carefully to achieve their best results.
The quantitative results of FID, KID, and LPIPS are summarized in Table~\ref{tab:FaceSRcomparison} (2-nd row).
% , proving the advanced performance of it in face SR
It can be seen that our SCGAN obtains higher indices than the other competiting methods.
The qualitative results of visual quality are presented in Figure~\ref{fig:fsrandrfsr} (left part).
One can see that DFDNet~\cite{DFDNet_eccv2018} and HifaceGAN~\cite{hifacegan_acmmm2020} produce blurry results similar to those produced by Bicubic Interpolation.
Real-ESRGAN, CycleGAN, and LRGAN fail to preseve well the facial structure.
We also observe that, although PULSE and GFPGAN have amazing generalization ability in face super-resolution, their performance still has room for improvement for very low-resolution face images ($16\times16$) that suffer from severe degradation.
On the contrary, our SCGAN generates realistic results in ensuring the structure and detail consistency to that of the ground-truth HR face images.

\noindent
\textbf{Face SR on \textsl{complex} degradation}.
Before generalizing our SCGAN to real-world face SR, we perform experiments on blind face SR with random \textsl{complex} degradation.
Here, the \textsl{complex} degradation with random blur kernel, downsampling, synthetic noise, and JPEG compression is performed on the 2,500 face images in FFHQ as the test set~\cite{StyleGAN_cvpr2019}, as described in \S\ref{sec:dataset}.
The quantitative results are listed in Table~\ref{tab:FaceSRcomparison} (3-rd row).
It can be seen that our SCGAN obtains clearly lower scores of FID, KID and LPIPS than the other methods.
In Figure~\ref{fig:fsrandrfsr} (right part), we compare the face SR results of different methods on representative face samples in FFHQ~\cite{StyleGAN_cvpr2019}.
We observe that Bicubic Interpolation, DFDNet~\cite{DFDNet_eccv2018} and HifaceGAN~\cite{hifacegan_acmmm2020} still produce blurry results.
Besides, Real-ESRGAN~\cite{realesrgan_cvpr2021} and CycleGAN~\cite{cyclegan2017} fail to recover the face contexts.
LRGAN~\cite{LRGAN} tends to produce incomplete face structure, and PULSE~\cite{pulse_cvpr2020} is prone to lose the identity information or important face components like eyeglasses, while GFPGAN~\cite{GFPGAN2021} fails to recover important face details.
On the contrary, our SCGAN generates high-quality and realistic HR face images with accurate face structure and fine-grained face details.
All these results validate that our SCGAN is more robust to the complex random degradation, and can produce high-quality HR face images more realistically to the real-world HR face images, than all the comparison methods.

\begin{table}[htbp]
  \centering
  \footnotesize
  \renewcommand{\arraystretch}{1.2}
  \setlength\tabcolsep{2.4pt}
  \caption{\small \textbf{Accuracy ($\%$) of face detection on the SR results restored by our SCGAN and other methods on two synthetic and two real-world datasets}.
  The best, second best, and third best results are highlighted in \textbf{\textcolor{red}{red}}, \textbf{\textcolor{blue}{blue}} and \textbf{bold}, respectively.}
  \vspace{-3mm}
\begin{tabular}{c||c|c|c|c}
  \Xhline{1.6pt}
Method &
  \begin{tabular}[c]{@{}c@{}}LS3D-W\\ balanced~\cite{LS3D_2017}\end{tabular} &
  \begin{tabular}[c]{@{}c@{}}FFHQ\\ ~\cite{StyleGAN_cvpr2019}\end{tabular} &
  \begin{tabular}[c]{@{}c@{}}Widerface\\ ~\cite{widerface_cvpr2016}\end{tabular} &
  WebFace \\   \hline\hline
Bicubic Interpolation & 42.80 & 42.04 & 53.90 & 46.49 \\ 
DFDNet~\cite{DFDNet_eccv2018}  & 32.90 & 31.64 & 41.90 & 35.99 \\ 
HifaceGAN~\cite{hifacegan_acmmm2020} & 43.10 & 41.36 & 42.05 & 45.53 \\ 
CycleGAN~\cite{cyclegan2017} & \textbf{84.90} & \textbf{86.92} & \textbf{\textcolor{blue}{87.90}} & \textbf{\textcolor{blue}{86.18}} \\ 
LRGAN~\cite{LRGAN} 
& 75.60 & 67.56 & \textbf{83.90} & \textbf{84.63} \\ 
GFPGAN~\cite{GFPGAN2021}                                                         & \textbf{\textcolor{blue}{92.10}} & \textbf{\textcolor{blue}{91.76}} & 83.55 & 74.90 \\ 
SCGAN (Ours) 
& \textbf{\textcolor{red}{96.60}} & \textbf{\textcolor{red}{95.40}} & \textbf{\textcolor{red}{97.75}} & \textbf{\textcolor{red}{96.89}} \\ \hline
\end{tabular}
\label{tab:detection}
\vspace{-3mm}
\end{table}

\noindent
\textbf{Face SR on \textsl{real-world} degradation}.
Now we compare different methods on the Widerface~\cite{widerface_cvpr2016}, our Webface and DroneSURF~\cite{dronesurf_2019} dataset for real-world face SR with complex and unknown degradation, where the experimental settings are described in \S\ref{sec:dataset}.
The quantitative results are presented in Table~\ref{tab:FaceSRcomparison}.

One can see that our SCGAN achieves higher FID and KID results than the other methods on Widerface and our WebFace datasets.
At the same time, it is only inferior to PULSE on DroneSURF ~\cite{dronesurf_2019} dataset.
As shown in Figure~\ref{fig:fsrandrfsr}, though achieving the best NIQE results in the Widerface dataset, HifaceGAN is prone to produce blurry face images, similar to Bicubic Interpolation, DFDNet, and Real-ESRGAN.
Though PULSE achieves the best FID and KID scores on DroneSURF dataset, its results produce noticeable structural changes from the input LR face images.
Although Real-ESRGAN achieves the best NIQE score on DroneSURF dataset, it is difficult to well recover some key facial features such as eyes, nose, etc.
The methods of LRGAN, PULSE, and GFPGAN produce either artifacts or color bias.
After all, our SCGAN not only restores the facial structure and details, but also preserves human identity of real-world LR face images, when compared to the other comparison methods.

\vspace{-3mm}

\subsection{Application on Downstream Vision Tasks}
\label{sec:application}

In this section, we apply our SCGAN and state-of-the-art methods to downstream vision tasks. We conduct experiment on face detection, face verification and face landmark detection in \S\ref{sec:facedetection}, \S\ref{sec:facerecognition} and \S\ref{sec:facelandmarkdetection}, respectively.
On all tasks, we compare our SCGAN with the methods of Bicubic Interpolation, DFDNet~\cite{DFDNet_eccv2018}, HifaceGAN~\cite{hifacegan_acmmm2020}, CycleGAN~\cite{cyclegan2017}, LRGAN~\cite{LRGAN}, and GFPGAN~\cite{GFPGAN2021} on face SR.

\begin{table}[t]
\centering
\caption{\small \textbf{Accuracy of face verification by FaceNet on the super-resolved face images in DroneSURF test set~\cite{dronesurf_2019} restored by different face super-resolution methods}.
The best, second best, and third best results are highlighted in \textbf{\textcolor{red}{red}}, \textbf{\textcolor{blue}{blue}} and \textbf{bold}, respectively.}
\begin{tabular}{r||c}
  \Xhline{1.6pt}
Method                & Accuracy ($\%$)                                            \\ \hline\hline
Bicubic Interpolation & 37.67                                               \\ \hline
DFDNet~\cite{DFDNet_eccv2018}                & 27.83
    \\ \hline
HifaceGAN~\cite{hifacegan_acmmm2020}             & \textbf{40.33}                                               \\ \hline
CycleGAN~\cite{cyclegan2017}              & 18.17                                                   \\ \hline
LRGAN~\cite{LRGAN}                 &  \textbf{\textcolor{blue}{47.83}}                                                   \\ \hline
GFPGAN~\cite{GFPGAN2021}                &  46.17                                                   \\ \hline
SCGAN(Ours)           &  \textbf{\textcolor{red}{56.83}}                                                 \\ \hline
\end{tabular}
\label{tab:recognition}
\end{table}

\subsubsection{Application on Face Detection}
\label{sec:facedetection}
Face detection is to predict the bounding boxs around the faces in the images.
To validate the effectiveness of these methods on face SR, we perform face detection with the state-of-the-art face detection method of RetinaFace~\cite{retinaface_CVPR2020},  on the SR face images by different methods.
% simple pre-trained model of~\cite{HOGdetection}, which is based on Histogram of Oriented Gradient (HOG)~\cite{HOG_CVPR2005} and Support Vector Machine (SVM)~\cite{SVM_1998}, on the SR face images by different methods.
% 
Here, we define face detection accuracy as the ratio of the number of face images successfully predicted by bounding boxes to the total number of the input face images, and each input image has exactly one face.
In Table~\ref{tab:detection}, we list the detection accuracies on the SR face images by different methods from the datasets of LS3D-W balanced~\cite{LS3D_2017}, FFHQ~\cite{StyleGAN_cvpr2019}, Widerface~\cite{widerface_cvpr2016} and WebFace, as described in \S\ref{sec:dataset}.
One can see that, the model of RetinaFace~\cite{retinaface_CVPR2020} consistently achieves the highest detection accuracy on the SR face images by our SCGAN.
This validates the effectiveness of our SCGAN on the face structure preservation for face detection.

\CheckRmv{
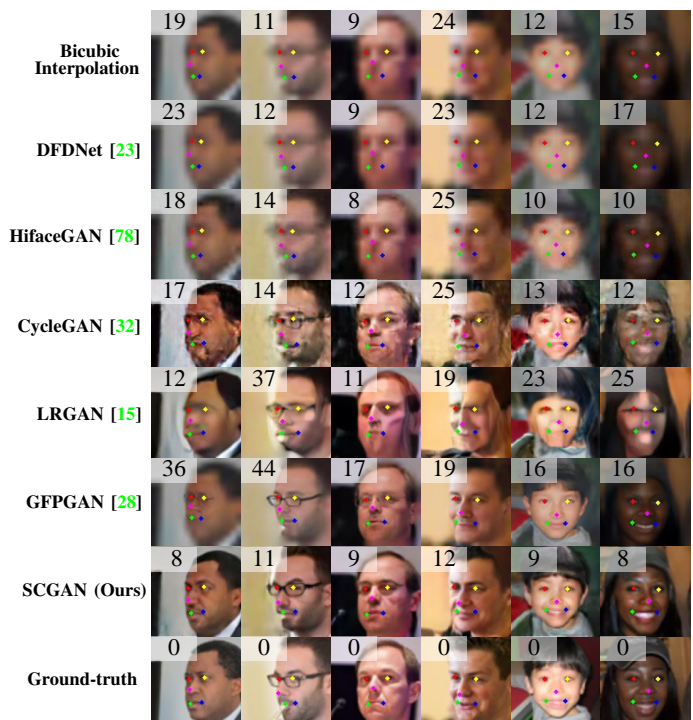
\begin{figure}[htbp] % htbp
	\centering
    \small
	\begin{overpic}[width=\linewidth]{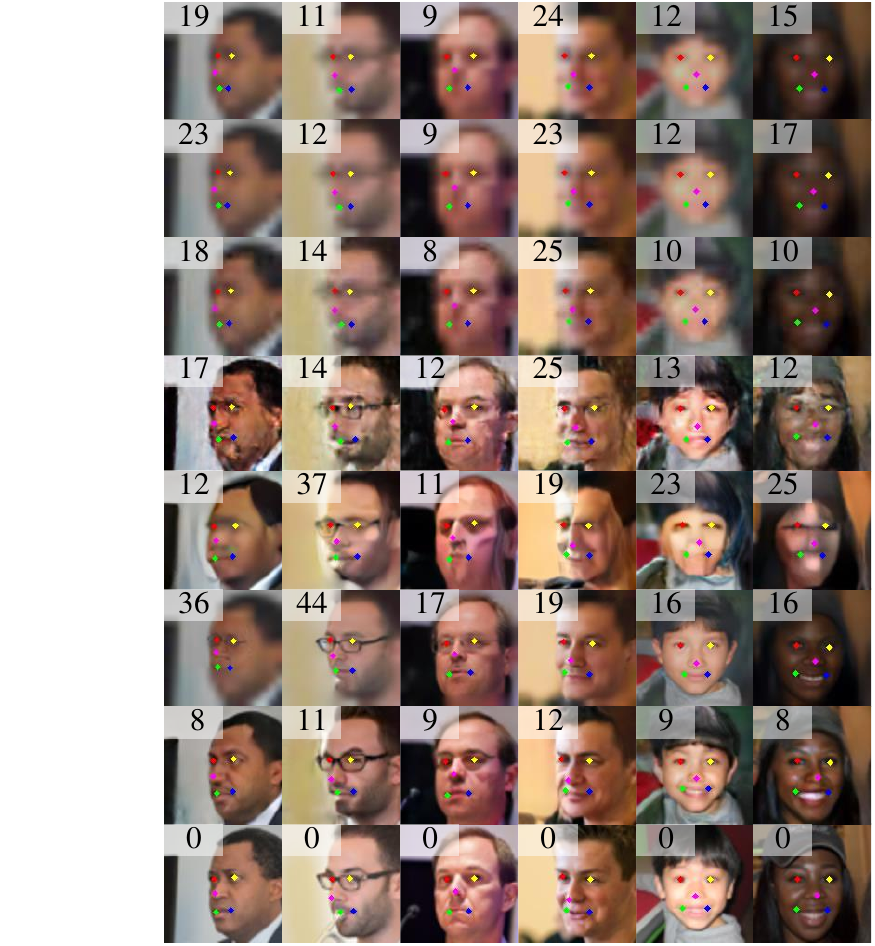}
    \put(4.7,93.3){\textbf{\scriptsize{Bicubic}}}
	\put(1.2,91.0){\textbf{\scriptsize{Interpolation}}}
	\put(1.5,79.6){\textbf{\scriptsize{DFDNet~\cite{DFDNet_eccv2018}}}}
	\put(-2.3,67.6){\textbf{\scriptsize{HifaceGAN~\cite{hifacegan_acmmm2020}}}}
	\put(-1.2,55.1){\textbf{\scriptsize{CycleGAN~\cite{cyclegan2017}}}}
	\put(1.5,42.9){\textbf{\scriptsize{LRGAN~\cite{LRGAN}}}}
	\put(0.0,30.7){\textbf{\scriptsize{GFPGAN~\cite{GFPGAN2021}}}}
    \put(-0.5,18.5){\textbf{\scriptsize{SCGAN (Ours)}}}
    \put(0.2,6.2){\textbf{\scriptsize{Ground-truth}}}
    \end{overpic}
    \vspace{-8mm}
\caption{\textbf{Comparison results of landmark detection and the corresponding $\ell_1$ norm errors} on the face images in synthetic FFHQ~\cite{StyleGAN_cvpr2019} dataset restored by different methods. Please zoom in for better view.
}
    \label{fig:landmarkdetection}
    \vspace{-2mm}
\end{figure}
}

\subsubsection{Application on Face Verification}
\label{sec:facerecognition}
Face verification is a binary classification task to determine whether the pair of output and reference face images have the same identity or not.
Here, we first restore the synthetic DroneSURF test set~\cite{dronesurf_2019}, as introduced in \S\ref{sec:dataset}, by different face SR methods.
%
% The SR face images by different methods are shown in Figure~\ref{fig:recognition}.
%
Then we perform face verification on the restored face images by the state-of-the-art face verification method of FaceNet~\cite{facenet_CVPR2015}.
The accuracies of FaceNet on the SR images by different methods are listed in Table~\ref{tab:recognition}.
It can be seen that, the accuracy on the SR images of our SCGAN is clearly higher than those of the other face SR methods.
This demonstrates the superiority of our SCGAN over the other competitors on preserving the consistency of face identity information for the face SR task.
% 
% One can see that the results obtained by our SCGAN maintain better the identity consistency than the other methods.
%
% Then we perform face verification on the restored face images by the state-of-the-art face verification method of FaceNet~\cite{facenet_CVPR2015}.
% %
% % For the calculation formula of VAL, please refer to~\cite{facenet_CVPR2015}.
% The accuracies of FaceNet on the SR results by different methods are listed in Table~\ref{tab:recognition}.
% %
% It can be seen that, except for the ground-truths, the accuracy on the SR results of our SCGAN is clearly higher than those of the other methods.
% %
% This demonstrates the superiority of our SCGAN over the other competitors on preserving the consistency of face identity information for the face SR task.

\subsubsection{Application on Face Landmark Detection}
\label{sec:facelandmarkdetection}
Face landmark detection aims to locate the key facial components of face images.
We restore the LR images into HR ones with more facial details and then use the state-of-the-art face landmark detection method of RetinaFace~\cite{retinaface_CVPR2020} for face landmarks detection.
We compare the face SR methods on six representative LR face images degraded from FFHQ test set, as described in \S\ref{sec:dataset}.
The landmark detection results of six representative face images restored by different methods and the $\ell_1$ norm errors (lower is better) between them and the corresponding ground-truth landmarks are shown in Figure~\ref{fig:landmarkdetection}.
% The landmark detection results on six representative face images restored by different methods are shown in Figure~\ref{fig:landmarkdetection}.
%
We observe that, compared with the SR results of other methods, the landmarks detected by the face images restored by our SCGAN are closer to those detected on the original HR face images, and the corresponding $\ell_1$ errors are also the lowest among all comparison methods.
This shows that our SCGAN recovers the key facial components more detectable than the other methods for face SR.

\section{Conclusion}
% and Future Work
\label{sec:conclusion}

In this paper, we proposed a novel Semi-Cycled Generative Adversarial Network (SCGAN) to alleviate the domain gap between unpaired LR and HR face images for real-world face super-resolution (SR).
% , which are coupled by a restoration branch,
Our SCGAN contains three independent branches to learn the forward and backward cycle-consistent reconstruction processes.
Specifically, a synthetic degradation branch learns to generate synthetic LR face images by degrading the real-world HR ones, a restoration branch recovers SR face images from the synthetic/real-world LR face images, and a real-world degradation branch degrades the SR face images restored from the real-world LR ones.
% both the forward and backward cycle-consistent learning processes
The restoration branch is coupled and regularized by the two independent degradation branches, making our SCGAN robust to super-resolve synthetic and real-world LR face images.
Experiments on two synthetic and two real-world datasets demonstrated that our semi-cycled SCGAN outperforms the state-of-the-art methods on synthetic and real-world face SR tasks, in terms of structure preservation, detail recovery, and standard objective metrics.
Three downstream vision tasks on face detection, face verification, and face landmark detection reveal that the effectiveness of our SCGAN better recovers the face structure, identity, and details than the other face SR methods.

% Although our SCGAN has achieved satisfactory performance in real-world face SR, it still faces some challenges. In some severe real-world scenarios (such as extremely dark, hazing, rainy, \etc.), designing a model that incorporates relevant priors may be a solution. Besides, how to use less information to perform restoration well, such as super-resolve some LR face images with a resolution of only $8\times8$ or even $4\times$4, etc., is also our interest. We have also found that, for some LR face images with uneven illumination, the results produced by our SCGAN may ignore this phenomenon and thus look unnatural. How to deal with these challenges more effectively will be our future work.

%\section*{Acknowledgments}
%This research is supported in part by The National Natural Science Foundation of China (No. 62002176, 62176068, and 61872225), Natural Science Foundation of Shandong Province (No. ZR2020MF038, ZR2020KF013, and ZR2020ZD44).

\bibliographystyle{IEEEtran}
\bibliography{FaceSR_final}

\vspace{-10mm}
\begin{IEEEbiography}
[{\includegraphics[width=1in,height=1.25in,clip,keepaspectratio]{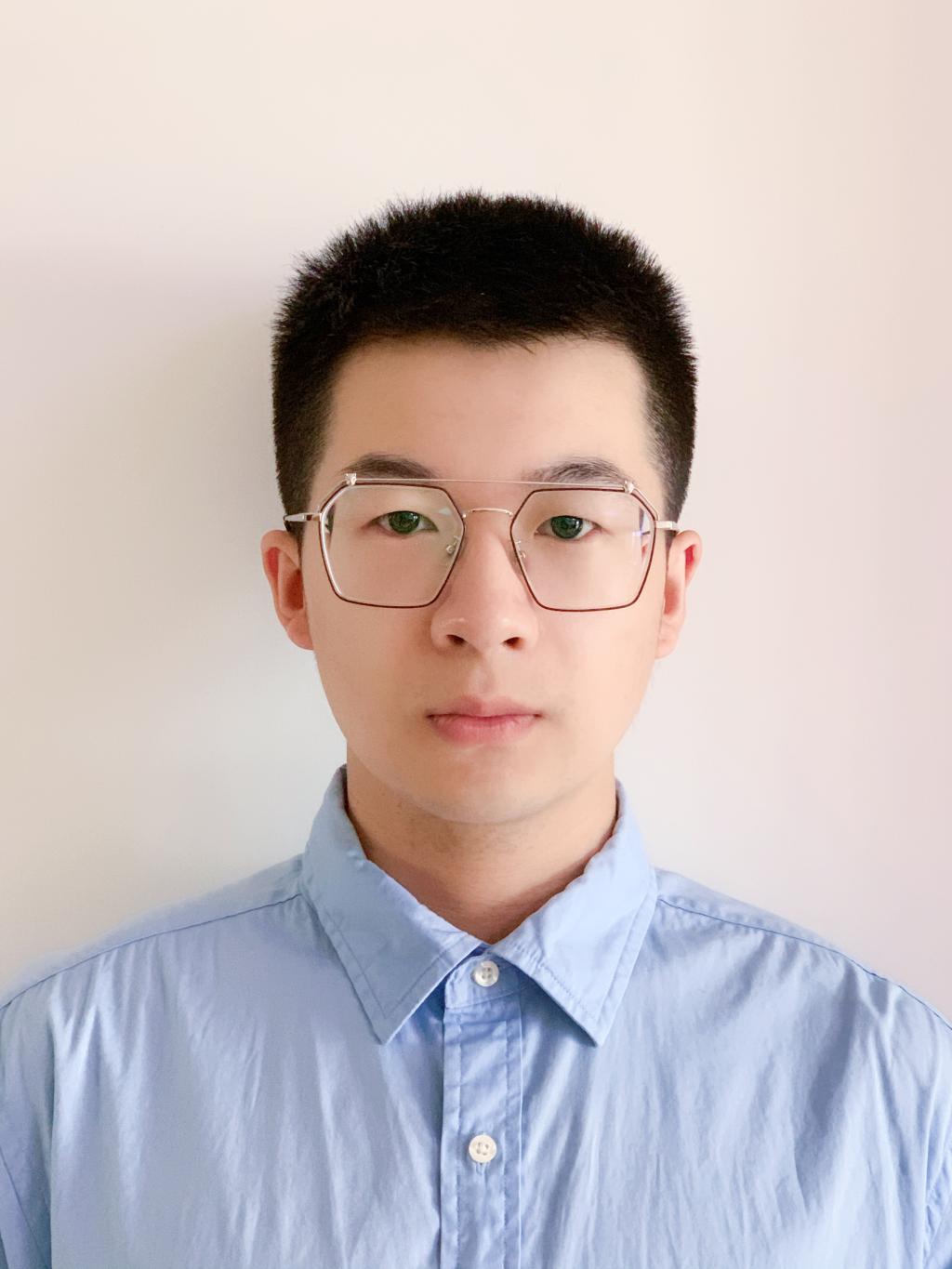}}]
{Hao Hou} received his B.Sc. degree in Information and Computing Science from the School of Mathematical Science, University of Jinan in 2019. He is currently an M.Sc. student with the College of Intelligence and Information Engineering, Shandong University of Traditional Chinese Medicine, Ji'nan, China. His current research interests are in the areas of image processing and medical image analysis.
\end{IEEEbiography}

\vspace{-12mm}
\begin{IEEEbiography}
[{\includegraphics[width=1in,height=1.25in,clip,keepaspectratio]{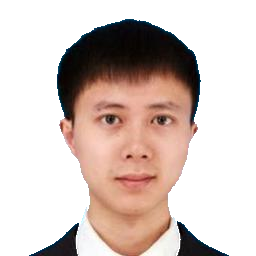}}]
{Jun Xu} received his B.Sc. and M.Sc. degrees from School of Mathematics Science, Nankai University, Tianjin, China, and his Ph.D. degree from the Department of Computing, Hong Kong Polytechnic University, in 2018. He worked as a Research Scientist at IIAI, Abu Dhabi, UAE. He is an Associate Professor with School of Statistics and Data Science, Nankai University. More information can be found at \url{https://csjunxu.github.io/}.
\end{IEEEbiography}

\vspace{-12mm}
\begin{IEEEbiography}
[{\includegraphics[width=1in,height=1.25in,clip,keepaspectratio]{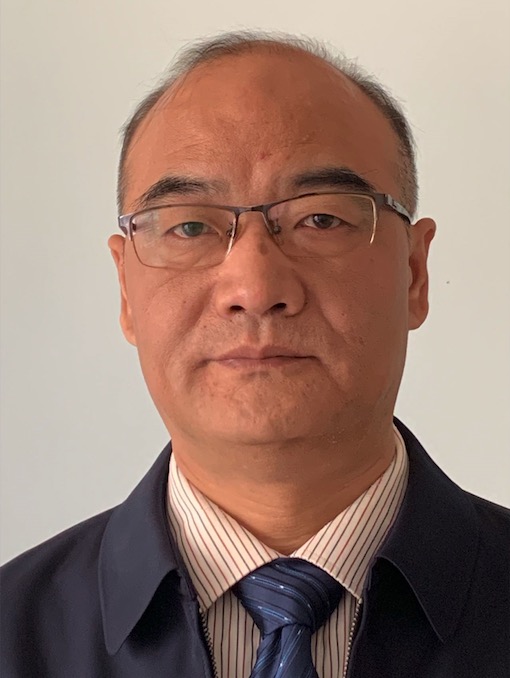}}]
{Yingkun Hou} received his Ph.D. degree from the School of Computer Science and Technology, Nanjing University of Science and Technology in 2012. He is currently a professor with the School of Information Science and Technology, Taishan University, Taian, China. His current research interests are in the areas of image processing, pattern recognition, and artificial intelligence.
\end{IEEEbiography}

\vspace{-12mm}
\begin{IEEEbiography}
[{\includegraphics[width=1in,height=1.25in,clip,keepaspectratio]{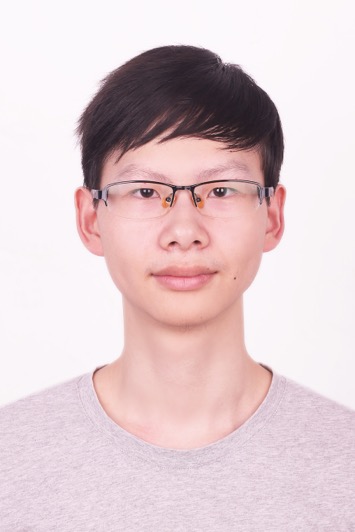}}]
{Xiaotao Hu} received his B.Sc. degree in School of Software Engineering, Dalian University of Technology in 2021. He is currently an M.Sc. student with the College of Computer Science, Nankai University, Tianjin, China. His current research interests are in the areas of image processing, medical image processing and analysis.
\end{IEEEbiography}

\vspace{-12mm}
\begin{IEEEbiography}
[{\includegraphics[width=1in,height=1.25in,clip,keepaspectratio]{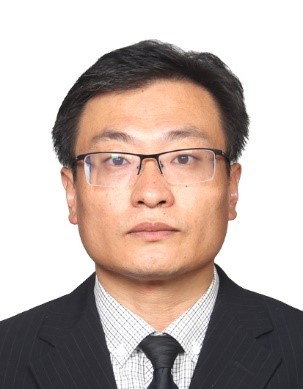}}]
{Benzheng Wei} received the B.S. degree in computer science from School of Computer Science at Shandong Institute of Light Industry, Jinan, China, in 2000, the M.S. degree in computer science from School of Computer Science and Technology at Shandong University, Jinan, China, in 2007, and the Ph.D. degree in precision instrument and machinery from College of Automation Engineering at Nanjing University of Aeronautics and Astronautics, Nanjing, China, in 2013. He is a professor with Shandong University of Traditional Chinese Medicine. He is also acting as a director at both the Center for Medical Artificial Intelligence and the Computational Medicine Lab of Shandong University of Traditional Chinese Medicine. His current research interests are in artificial intelligence, medical information engineering and computational medicine. He has published over 80 papers in refereed international leading journals/conferences such as Medical Image Analysis, IEEE TMI, Neurocomputing, IPMI and MICCAI.
\end{IEEEbiography}

\vspace{-10mm}
\begin{IEEEbiography}
[{\includegraphics[width=1in,height=1.25in,clip,keepaspectratio]{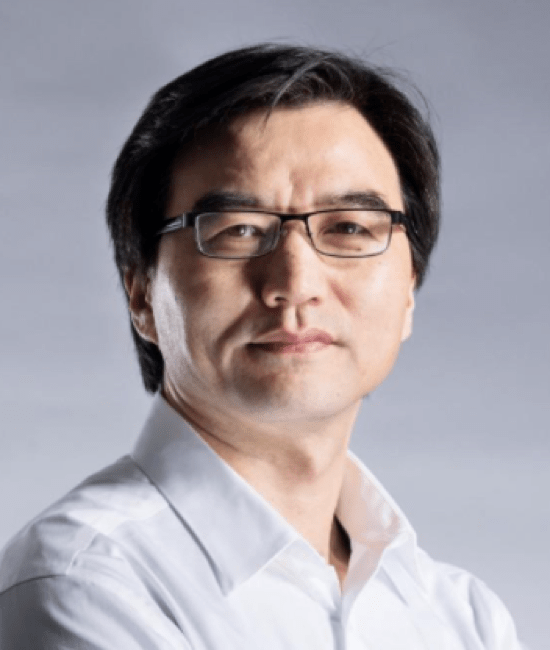}}]
{Dinggang Shen} received the Ph.D. degree in electrical engineering from Shanghai Jiao Tong University, Shanghai, China, in 1995. He has published more than 1100 peer-reviewed papers in the international journals and conference proceedings, with H-index 127. His research interests include medical image analysis, computer vision, and pattern recognition. Dr. Shen served on the Board of Directors, the Medical Image Computing and Computer Assisted Intervention (MICCAI) Society, in 2012–2015, and was the General Chair for MICCAI 2019. He is a fellow serves as an editorial board member for eight international journals. He is a Fellow of the American Institute for Medical and Biological Engineering and the International Association for Pattern Recognition. He is currently the founding dean of the School of Biomedical Engineering, ShanghaiTech University, Shanghai, China.
\end{IEEEbiography}

\vfill

\end{document}